\documentclass[letterpaper, 10 pt, conference]{ieeeconf}
\usepackage{amsmath,amsfonts}
\usepackage{algorithmic}
\usepackage{array}
\usepackage{textcomp}
\usepackage{stfloats}
\usepackage{verbatim}
\usepackage{graphicx}
\usepackage{balance}
\usepackage{amssymb}
\usepackage{epsfig}
\usepackage{multirow}
\usepackage{tabularx}
\usepackage{adjustbox}
\usepackage{siunitx}
\usepackage{makecell}
\usepackage{stmaryrd} 
\usepackage{latexsym} 
\usepackage{url}
\usepackage{amsfonts}
\usepackage{xkeyval}

\usepackage[breaklinks,colorlinks,bookmarks]{hyperref}
\usepackage{booktabs}

\usepackage{subcaption}

\usepackage{xcolor,colortbl}
\definecolor{Gray1}{gray}{0.85}
\definecolor{Gray2}{gray}{0.65}
\definecolor{darkgreen}{RGB}{0,127,0}
\definecolor{darkred}{RGB}{200,0,0}

\usepackage{pifont}
\usepackage[capitalize]{cleveref}
\crefname{section}{Sec.}{Secs.}
\Crefname{section}{Section}{Sections}
\Crefname{table}{Table}{Tables}
\crefname{table}{Tab.}{Tabs.}

\newcommand{\figref}[1]{Fig.~\ref{#1}}

\newcommand{\Tabref}[1]{Table~\ref{#1}}

\newcommand{\ie}{\textit{i.e.}}

\newcommand{\cf}{\textit{cf.}}

\definecolor{Aquamarine}{rgb}{0.32, 0.70, 0.73}

\usepackage{bbm}

\IEEEoverridecommandlockouts
\usepackage{graphics}
\usepackage{epsfig}
\usepackage{mathptmx}
\usepackage{times}
\usepackage{amsmath}
\usepackage{amssymb}
\usepackage{arydshln}
\usepackage{cite}

\title{\LARGE \bf
MrGS: Multi-modal Radiance Fields with 3D Gaussian Splatting for RGB-Thermal Novel View Synthesis
}

\author{Minseong Kweon$^{1^{\dagger}}$, Janghyun Kim$^{2^{\dagger}}$, Ukcheol Shin$^{3}$, and Jinsun Park$^{4*}$
\thanks{$^{1}$Minseong Kweon is with the Minnesota Robotics Institute (MnRI),
University of Minnesota, Twin Cities, MN 55455, USA
(e-mail: kweon021@umn.edu).}
\thanks{$^{2}$Janghyun Kim is with the Department of Information Convergence Engineering (Artificial Intelligence Major), Pusan National University, Busan, Republic of Korea (e-mail: jangjoa41@pusan.ac.kr).}
\thanks{$^{3}$Ukcheol Shin is with the Department of Energy Engineering, Korea Institute of Energy Technology (KENTECH), Jeonnam, Republic of Korea (e-mail: ushin@kentech.ac.kr).} 
\thanks{$^{4}$Jinsun Park is with the School of Computer Science and Engineering, Pusan National University, Busan, Republic of Korea (e-mail: jspark@pusan.ac.kr).} 
\thanks{$^{\dagger}$ Equal contribution}
\thanks{$*$ Corresponding author}
}

\begin{document}

\maketitle

\begin{abstract}
Recent advances in Neural Radiance Fields (NeRFs) and 3D Gaussian Splatting (3DGS) have achieved considerable performance in RGB scene reconstruction. However, multi-modal rendering that incorporates thermal infrared imagery remains largely underexplored. Existing approaches tend to neglect distinctive thermal characteristics, such as heat conduction and the Lambertian property. In this study, we introduce MrGS, a multi-modal radiance field based on 3DGS that simultaneously reconstructs both RGB and thermal 3D scenes. Specifically, MrGS derives RGB- and thermal-related information from a single appearance feature through orthogonal feature extraction and employs view-dependent or view-independent embedding strategies depending on the degree of Lambertian reflectance exhibited by each modality. Furthermore, we leverage two physics-based principles to effectively model thermal-domain phenomena. First, we integrate Fourier's law of heat conduction prior to alpha blending to model intensity interpolation caused by thermal conduction between neighboring Gaussians. Second, we apply the Stefan-Boltzmann law and the inverse-square law to formulate a depth-aware thermal radiation map that imposes additional geometric constraints on thermal rendering. Experimental results demonstrate that the proposed MrGS achieves high-fidelity RGB-T scene reconstruction while reducing the number of Gaussians.
\end{abstract}

\section{Introduction}

\begin{figure}[t]
    \centering
    \begin{subfigure}[t]{1.0\linewidth}
        \centering
        \includegraphics[width=1.00\linewidth]{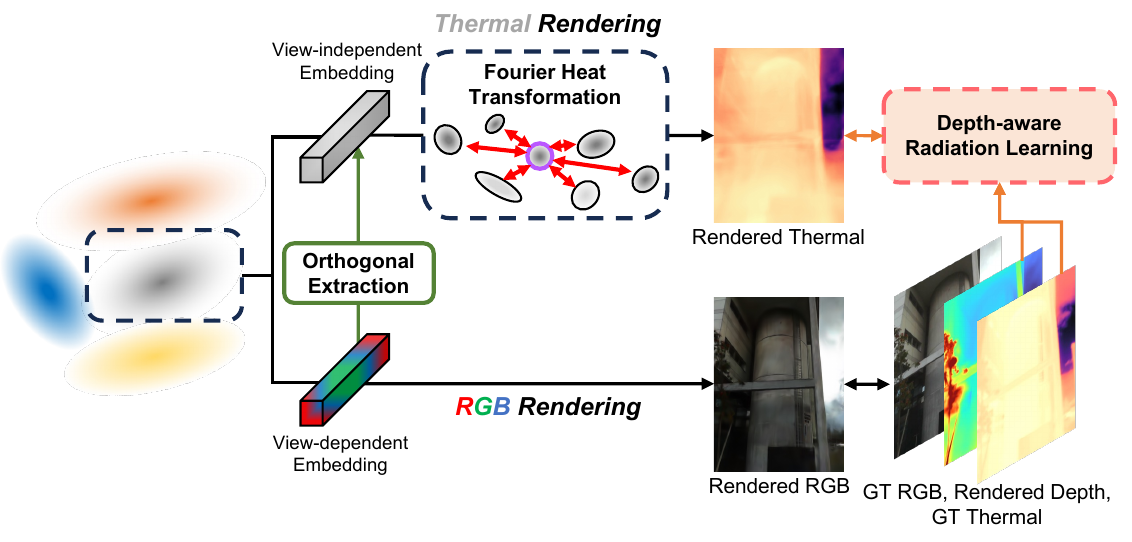} \\
        \caption{\footnotesize Overview of our proposed framework for RGB-T scene reconstruction.}
        \label{fig:teaser_arch}
    \end{subfigure}
    \par\smallskip
    \begin{subfigure}[t]{1.0\linewidth}
        \centering
        \renewcommand{\arraystretch}{0.3}
        \begin{tabular}{c@{\hskip 0.003\linewidth}c@{\hskip 0.003\linewidth}c@{\hskip 0.003\linewidth}c}
        \includegraphics[width=0.245\linewidth]{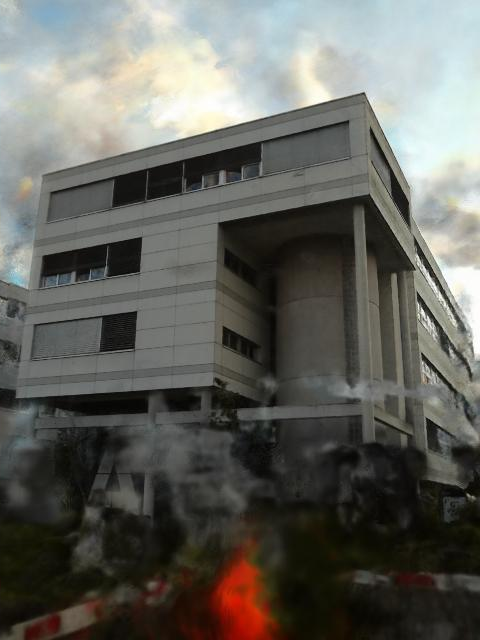}  & \includegraphics[width=0.245\linewidth]{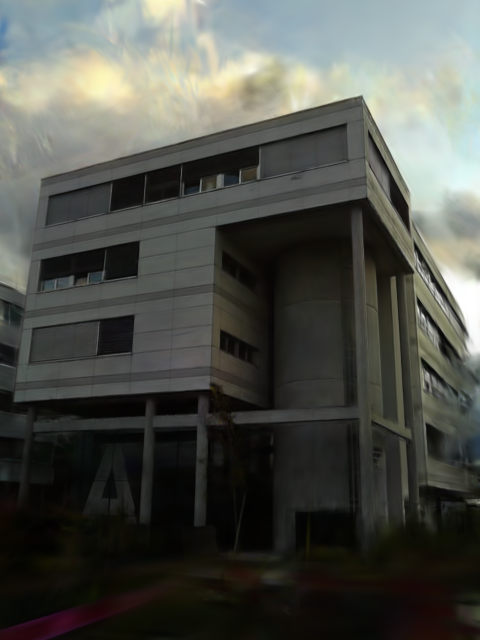} &
        \includegraphics[width=0.245\linewidth]{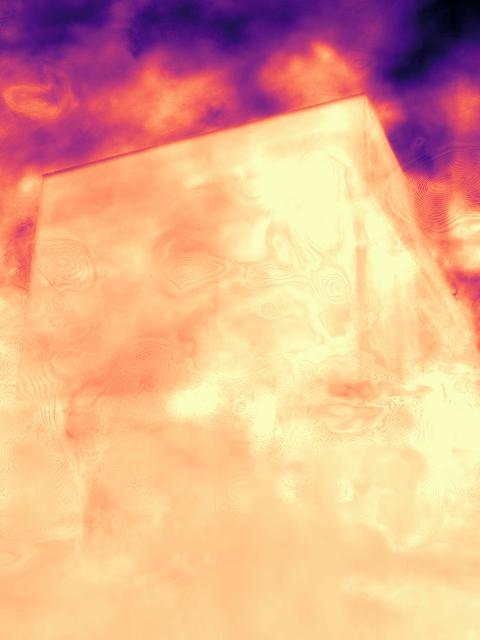}  & 
        \includegraphics[width=0.245\linewidth]{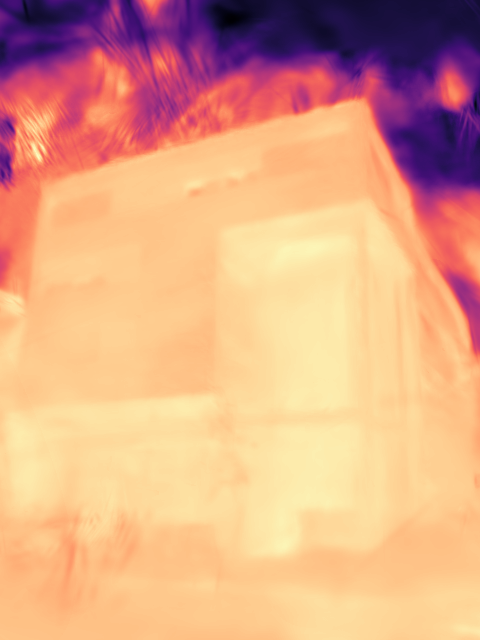} \\
    \footnotesize ThermoNeRF~\cite{hassan2024thermonerf} & \footnotesize MrGS (ours) & \footnotesize ThermoNeRF~\cite{hassan2024thermonerf} & \footnotesize MrGS (Ours) \\
    \end{tabular}
    \caption{\footnotesize Comparison between ThermoNeRF~\cite{hassan2024thermonerf} and our proposed MrGS.}
        \label{fig:teaser_pred_comparison}
    \end{subfigure}
    \caption{\textbf{Overview of our MrGS and comparisons of RGB-T scene rendering results with a NeRF-based method.} 
    }
    \label{fig:teaser_compare}
\end{figure}
Novel view synthesis has become a key technology for 3D reconstruction in real-world applications such as augmented reality (AR), virtual reality (VR), and autonomous driving.
Various methods have been explored for scene reconstruction, such as Neural Radiance Fields (NeRFs)~\cite{mildenhall2021nerf} and 3D Gaussian Splatting (3DGS)~\cite{kerbl20233d}. 
Recently, thermal scene reconstruction has gained increasing attention, as RGB-based methods struggle in challenging scenarios such as wild environments or low-light conditions.
NeRF-based~\cite{xu2024leveraging} and 3DGS-based~\cite{chen2024thermal3dgs, liu2025thermalgs} methods have been introduced for 3D thermal scene rendering, while multi-modal NeRFs~\cite{hassan2024thermonerf, lin2024thermalnerf} aim to render RGB and thermal modalities jointly.
However, NeRF-based multi-modal methods struggle to reconstruct thermal scenes captured by infrared cameras because of the inherently textureless appearance of thermal imagery.

To address this issue, we introduce MrGS, a novel multi-modal radiance field framework based on 3DGS that leverages modality-specific characteristics to simultaneously render high-fidelity RGB and thermal images, as illustrated in \cref{fig:teaser_arch}.
Specifically, MrGS employs a multi-modal appearance embedding to encompass the characteristics of two independent modalities (\ie, color and temperature) for the same 3D scene within a single embedding.
After that, we consider view-independent reflection in thermal modality and view-dependent reflection in RGB modality by introducing two positional embeddings, along with orthogonal feature extraction.
Additionally, we incorporate physics-driven principles to accurately model the behavior of the thermal domain: i) Fourier’s law of heat conduction~\cite{fourier} to model heat transfer between Gaussians, ii) the Stefan-Boltzmann law~\cite{stefan1879uber} and inverse-square laws~\cite{adelberger2003tests} to formulate a depth-aware thermal radiation map, enforcing additional geometric constraints for thermal rendering. As shown in \cref{fig:teaser_pred_comparison}, our method demonstrates superior reconstruction quality compared to existing NeRF-based approaches.
Our contributions can be summarized as follows:

\begin{itemize}
\item We introduce MrGS, a novel multi-modal radiance field framework based on 3DGS for the simultaneous high-fidelity reconstruction of RGB and thermal images.
\item We propose a multi-modal appearance embedding and orthogonal feature extraction that effectively captures multi-modal characteristics.
\item We introduce Fourier’s law of heat conduction into thermal scene rendering to model real-world thermal flow within 3D Gaussians, enabling more accurate thermal scene reconstruction.
\item We propose depth-aware thermal radiation learning using the Stefan-Boltzmann law and inverse-square law to improve geometry awareness in thermal scene reconstruction.
\end{itemize}

\section{Related Work}
\label{sec:related_work}

\subsection{Novel View Synthesis}
3D reconstruction is a core task in computer graphics and vision. Traditional methods like Structure-from-Motion (SfM)~\cite{schonberger2016structure} and Multi-View Stereo (MVS)~\cite{yan2021deep} recover 3D geometry from images. With deep learning advancements, Neural Radiance Fields (NeRFs)~\cite{mildenhall2021nerf} have revolutionized novel view synthesis research by enabling photorealistic rendering and modeling view-dependent appearances.

More recently, 3D Gaussian Splatting (3DGS)~\cite{kerbl20233d} has emerged as a promising technique by representing scenes with anisotropic 3D Gaussians, pushing the boundaries of both rendering quality and efficiency. Its effectiveness has led to NeRFs being replaced across various 3D scene understanding tasks.
It is widely used in fields such as 3D content generation~\cite{tang2023dreamgaussian, liang2024luciddreamer, tang2024lgm, szymanowicz2024splatter}, mesh extraction~\cite{wu2024surface, choi2024meshgs, krishnan20253d, guedon2024sugar}, 3D semantic segmentation~\cite{zhou2024feature, jurca2024rt, ye2024gaussian, dou2024cosseggaussians}, and SLAM~\cite{yugay2023gaussian, zhu2024semgauss, li2024sgs, sun2024mm3dgs} to name a few.
One widely used approach is to incorporate geometric optimization through depth reconstruction, where rasterization extracts per-Gaussian depth from COLMAP~\cite{schonberger2016structure} outputs, providing depth constraints~\cite{chung2024depth, turkulainen2024dn, kerbl2024hierarchical} to integrate geometric information.
As surface normals derive from depth gradients, several studies~\cite{turkulainen2025ags, zhang20242dgs, guedon2024sugar, choi2024meshgs} use them as geometric priors to integrate mesh information.
3DGS’s point-based rendering naturally integrates with LiDAR, enabling a multi-modal framework that enhances geometric accuracy and rendering quality. Several studies~\cite{lim2024lidar, zhao2024tclc, xiao2024liv, sun2024mm3dgs, herau20243dgs} have combined LiDAR and RGB data to improve scene reconstruction, but they mainly focus on geometric fidelity, overlooking other modalities that could enhance scene understanding.
For instance, thermal imaging offers structural and material insights independent of illumination and texture, complementing RGB. However, existing multi-modal 3DGS frameworks overlook thermal data, limiting their applicability in robust environmental perception.

In this work, we introduce a multi-modal radiance field integrating thermal images with RGB within the 3DGS framework. By incorporating thermal information, our approach not only improves scene reconstruction but also enhances material awareness and structural consistency, demonstrating the potential of thermal imaging in 3D Gaussian-based rendering.

\subsection{3D Thermal Imaging}
Thermal images are widely used in optical and 3D vision applications due to their robustness to adverse weather, obscurants, and varying illumination.
One notable application is depth estimation~\cite{kim2018multispectral, lu2021alternative, shin2021self, kim2024exploiting}, which leverages thermal images in challenging outdoor and low-light environments, where conventional RGB sensors struggle.
Moreover, several studies~\cite{shin2019sparse, delaune2019thermal, saputra2021graph, wang2023edge, chen2023thermal, keil2024towards} have investigated thermal cameras for localization and SLAM.

Unlike these works, thermal scene reconstruction from infrared (IR) images faces challenges due to low contrast and lack of texture.
Therefore, previous studies have addressed this by integrating RGB images~\cite{ma2019visible} or pose information~\cite{lang20083d} to enrich geometric details.
With the emergence of NeRFs~\cite{mildenhall2021nerf} and 3DGS~\cite{kerbl20233d}, several methods have been proposed to reconstruct and process IR scenes. X-NeRF~\cite{poggi2022cross} optimizes cross-spectral camera poses to render aligned RGB and IR at the same resolution from arbitrary viewpoints. ThermoNeRF~\cite{hassan2024thermonerf} and ThermalNeRF~\cite{lin2024thermalnerf} propose a NeRF-based multi-modal approach for simultaneously rendering RGB and thermal views.
Thermal3D-GS~\cite{chen2024thermal3dgs} incorporates physical laws into 3DGS to model atmospheric transmission and thermal conduction, enabling effective thermal scene reconstruction. However, multi-modal integration of RGB and thermal data within 3DGS remains unexplored.

To make 3D Gaussians simultaneously render both RGB and thermal scenes, we propose a multi-modal 3DGS framework that considers the physical properties of two modalities.
For instance, we apply Fourier’s law of heat conduction to model thermal flow between 3D Gaussians for thermal images.
We also incorporate depth-aware thermal radiation learning by applying the Stefan-Boltzmann law and the inverse-square law.

\begin{figure*}
  \centering
  \includegraphics[width=\linewidth]{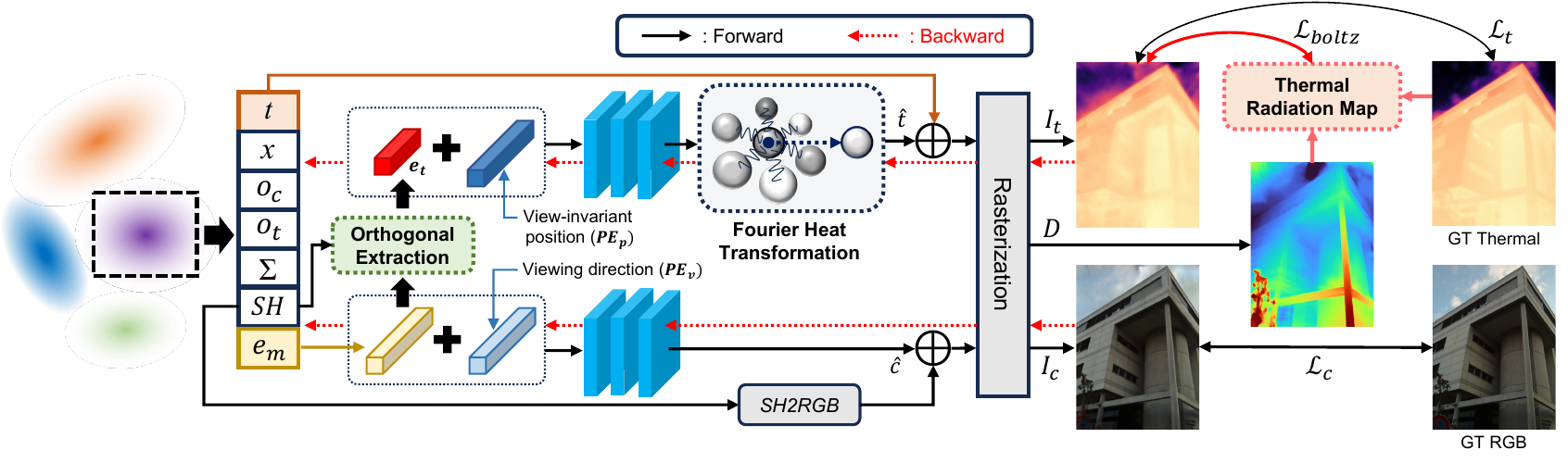}
  \caption{\textbf{Overall framework of the proposed MrGS.} 
  MrGS framework simultaneously renders RGB and thermal images from 3D Gaussians.
  The RGB branch integrates the multi-modal appearance embedding $e_m$ with view-dependent encoding $PE_v$ and SH2RGB conversion for rasterization. The thermal branch extracts the $e_{t}$ via orthogonal extraction, concatenates it with view-invariant encoding $PE_p$, and refines the features via Fourier heat transformation. Thermal image is further guided by a thermal radiation map with $L_{boltz}$.
  }
  \label{fig:main}
\end{figure*}

\section{Method}
\subsection{Preliminary}
3D Gaussian Splatting~\cite{kerbl20233d} represents an entire scene as a set of 3D Gaussians, each characterized by a center position $\mu$ that indicates its spatial location and a covariance matrix $\Sigma$ that determines its shape and orientation.
The opacity $o$ controls the transparency of the Gaussian, influencing how it blends with other elements in the scene.
Additionally, spherical harmonics $SH$ encode view-dependent radiance to model RGB appearance.
For a given position $x$, a single 3D Gaussian distribution is defined as:
\begin{equation}
G(x) = e^{-\frac{1}{2} (x - \mu)^T \Sigma^{-1} (x - \mu)}, \quad \Sigma = R S S^T R^T,
\end{equation}
where \( R \) is a rotation matrix that encodes the orientation of the Gaussian in 3D space, and \( S \) is a diagonal scaling matrix that defines the three principal axes.
$G(x)$ represents the spatial density function of a Gaussian, defining its influence at position $x$ based on the mean $\mu$ and covariance $\Sigma$, which determine its location, shape, and spread in 3D space.
These 3D Gaussians are then projected onto a 2D screen space, where a tile-based rasterizer sorts them and applies alpha-blending, as formulated below:
\begin{equation}
\alpha_i = \sigma(o_i)G'_i(x'),
\label{eq:pre-alphblending}
\end{equation}
\begin{equation}
I_c(x') = \sum_{i \in N}c_i\alpha_i\prod_{j=1}^{i-1} \left(1 - \alpha_j\right),
\label{eq:alphblending}
\end{equation}
where $G'_i$ and $c_i$ denote the projected 2D Gaussian and color of the $i$-th Gaussian, respectively.
$x'$ represents the pixel position in the screen space and $\sigma(\cdot)$ denotes the sigmoid function. N denotes the number of 2D Gaussians associated with $x'$.
This differentiable rasterization process allows the Gaussians' parameters to be optimized end-to-end during training.

\subsection{Multi-modal Radiance Fields with 3D Gaussian}
\subsubsection{Rendering Overview}
\Cref{fig:main} illustrates our proposed MrGS, a multi-modal radiance field framework with 3D Gaussian.
Our MrGS framework consists of two rendering branches for RGB and thermal images.
We assign separate opacity according to modality, $o_c$ for RGB scenes and $o_t$ for thermal scenes.

\noindent\textbf{RGB image rendering.} \ 
Given Gaussian parameters (\ie, $\mu, \Sigma, x, o_c, SH$), we additionally initialize multi-modal appearance embedding $e_{m}$ to encompass RGB scene characteristics.
After that, the multi-modal embedding is concatenated with the viewing direction embedding $PE_{v}(x)$ to capture view-dependent reflection in the RGB modality, and the combined feature is then fed through MLP layers to obtain color value $\hat{c}$ inferred from the multi-modal perspective.
In parallel, a SH2RGB module~\cite{kulhanek2024wildgaussians} transforms spherical harmonic (SH) coefficients into color value $c$ by considering lighting and material properties.
Lastly, the RGB image $I_{c}$ is rendered with $c$ and $\hat{c}$ using alpha blending as follows:
\begin{equation}
I_{c} = \sum_{i \in N}{(\hat{c}_{i} + c_{i})}\alpha_i\prod_{j=1}^{i-1} \left(1 - \alpha_j\right),
\label{eq:rgb_alphblending}
\end{equation}
where $\alpha$ is computed using \cref{eq:pre-alphblending} with $o_{c}$.

\noindent\textbf{Thermal image rendering.} \ 
In contrast to RGB image rendering, thermal image rendering must account for view-independence, as thermal radiation is invariant to viewing direction~\cite{hassan2024thermonerf}.
Therefore, we extract the view-invariant component $e_{t}$ from the multi-modal embedding $e_{m}$ using our proposed orthogonal extraction, which excludes the view-dependent components derived by $SH$ embedding.
Next, it is combined with view-invariant positional embedding $PE_{p}(x)$ and processed through MLP layers to generate feature $F_t$.
After that, our proposed Fourier heat transformation, which models heat transfer effects by considering interactions between neighboring Gaussians, further refines $F_t$ and estimates the thermal value $\hat{t}$ inferred from the multi-modal perspective.
Finally, the thermal image $I_t$ is computed with $t$ and $\hat{t}$ by applying alpha blending as follows:
\begin{equation}
I_{t} = \sum_{i \in N}{(\hat{t_{i}}+t_{i})}\alpha_i\prod_{j=1}^{i-1} \left(1 - \alpha_j\right),
\label{eq:thr_alphblending}
\end{equation}
where $\alpha$ is computed using \cref{eq:pre-alphblending} with $o_{t}$.

\noindent\textbf{Depth map rendering.} \ 
The depth map $D$ is estimated using a discrete volume rendering approximation, similar to \cref{eq:alphblending}, following the equation:
\begin{equation}
D = \sum_{i \in N}{{d_i}\alpha_i}\prod_{j=1}^{i-1} \left(1 - \alpha_j\right),
\label{eq:depth_map_rendering}
\end{equation}
where $d_i$ is the $z$-depth coordinate of the $i$-th Gaussian, and $\alpha$ is applied according to each modality.

\subsubsection{Orthogonal Embedding Extraction}
\begin{figure}
  \centering
  \includegraphics[width=0.85\linewidth]{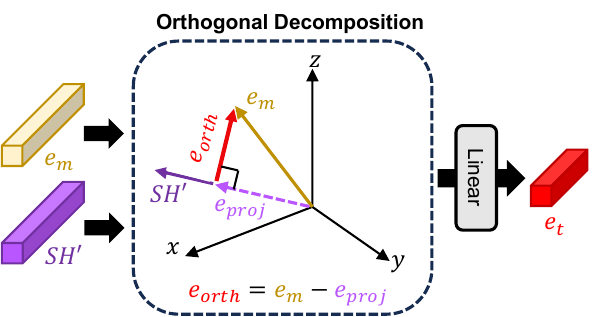}
  \caption{\textbf{Graphical description of the view-invariant component extraction for thermal image with orthogonal decomposition.}}
  \label{fig:ortho}
\end{figure}

By optimizing multi-modal radiance fields, we assume that the multi-modal appearance embedding $e_m$ encompasses both view-dependent and view-invariant characteristics for the same 3D scene.
Therefore, we aim to extract view-invariant component $e_{t}$ from $e_m$ for thermal image rendering.
Since Gaussian’s spherical harmonics primarily encode RGB lighting variations, we presume that excluding the related embedding from $e_m$ can produce view-invariant embedding only.  
To this end, we employ the orthogonal decomposition method~\cite{yang2021dolg} to disentangle independent information between the \(e_{m}\) and \(SH\), as shown in \cref{fig:ortho}, yielding the following formulation:
\begin{gather}
     e_{proj} = \frac{e_m\cdot SH'}{{\left | SH'\right |}^{2}} SH', \label{eq:embedding_proj} \quad SH' = \phi_{SH}(SH)\\
     e_{t} = \phi_{e}(e_{orth}) = \phi_{e}(e_{m} - e_{proj}), \label{eq:embedding_t}
\end{gather}
where $\phi_{SH}(\cdot)$ and $\phi_{e}(\cdot)$ denote single linear layers.

\subsubsection{Modeling Lambertian Reflectance}
Our pipeline is designed to handle both non-Lambertian (view-dependent) and Lambertian (view-independent) properties. While RGB images exhibit strong view dependency due to illumination variations and specular reflections, thermal imaging remains largely view-independent, as temperature measurements are stable across different viewpoints.
To integrate these characteristics into our multi-modal radiance field, we incorporate view-invariant positional embedding $PE_{\text{p}}(x)$ for thermal rendering and viewing direction embedding $PE_{\text{v}}(x)$ for RGB rendering as follows:
\begin{equation}
    PE(x) = \mathtt{concat} \left[ \sin(2^n\pi x), \cos(2^n\pi x)\right]_{n=0}^{L-1},
\label{eq:positional_embedding}
\end{equation}
where $2L$ is the encoding dimension, $\mathtt{concat}(\cdot)$ is the concatenation, and \(x\) represents the center position of the 3D Gaussian when rendering thermal images or the viewing direction when rendering RGB images.
Through this encoding, temperature values are learned based on spatial positions in 3D space, while color values are modeled according to the viewing direction.

\subsection{Fourier Heat Transformation}
\label{sec:fourier}

Since each Gaussian contributes to per-pixel thermal values, considering interactions between Gaussians is crucial.
Unlike color blending, thermal values spread across neighboring Gaussians, causing changes in their temperature.
Neglecting this interaction may result in unrealistic temperature distributions and visual artifacts.
\begin{figure}
  \centering
  \includegraphics[width=\linewidth]{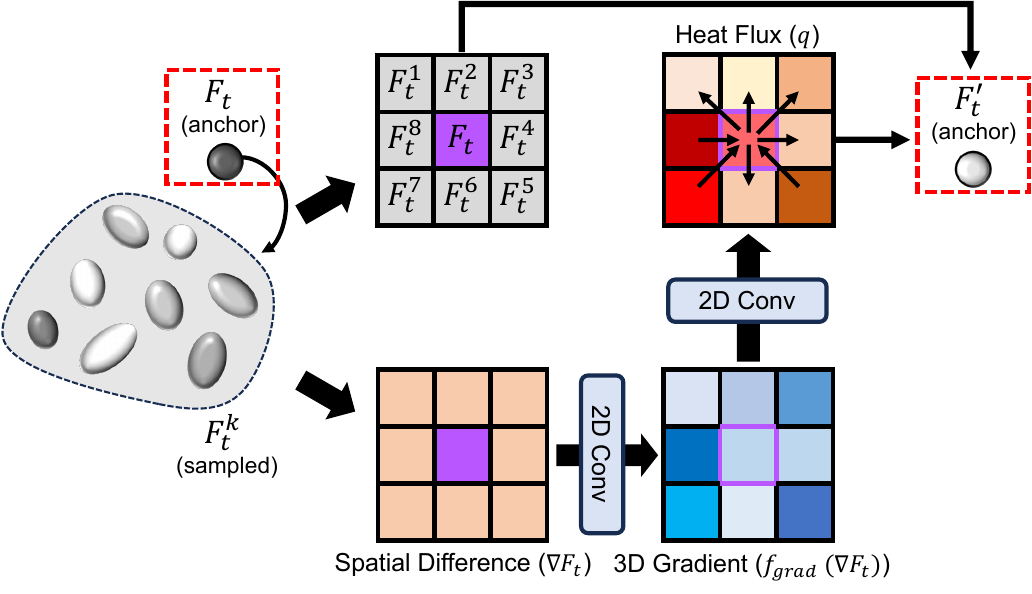}
  \caption{\textbf{Overall Process for refining thermal feature $F_t$ via Fourier's Heat Transformation.}
  }
  \label{fig:fourier}
\end{figure}
To address this, we propose the Fourier heat transformation, as illustrated in \cref{fig:fourier}.
This transformation applies Fourier’s law of heat conduction~\cite{fourier} to model regional heat transfer:
\begin{equation}
    q = -k \nabla t,
\end{equation}
where \(q\) represents the heat flux, \(k\) is the thermal conductivity coefficient, and \(\nabla t\) denotes the temperature difference, describing both the magnitude and direction.

By explicitly modeling heat transfer between Gaussians, our method mitigates temperature discontinuities and reduces rendering artifacts.
To integrate this into the 3DGS framework, we approximate Fourier’s Heat Conductive law as follows:
\begin{equation}
    q_{i} = f_{q}(-o_{t_{i}}\cdot f_{grad}(\nabla F_{t_{i}})),
    \label{eqs:our_fourier}
\end{equation}
where \( q_{i} \) denotes the heat flux, \( \nabla F_{t_{i}} \) represents the thermal feature's spatial gradient between the \( i \)-th Gaussian and its neighboring samples, \( o_{t_{i}} \) is its opacity, and \( f_{q} \) and \( f_{{grad}} \) are single-layer convolutions trained to approximate heat flux and gradients, respectively.
In specific, \( \nabla F_{t_{i}} \) is obtained by performing a K-Nearest Neighbors ($K$-NN) search on 3D Gaussians, where \( K \) Gaussians are sampled for each anchor Gaussian to index neighboring Gaussians.
Note that we apply Fourier's Heat Conductive law to the thermal feature $F_{t}$ to enable interaction between pixels in a high-level feature space.
Lastly, the refined thermal feature $F'_{t_{i}}$ of the anchor Gaussian is obtained as follows:
\begin{equation}
F'_{t_{i}} = w_{a} \left(F_{t_{i}} + q_{i}\right) + w_{s} \frac{\sum_{u \in U} F_{t_{u}}}{|U|},
\label{eq:final_thermal_feature}
\end{equation}
where $w_{a}$ and $w_{s}$ are weighting factors for the anchor and surrounding Gaussians, respectively. 
$F_{t_{u}}$ represents the thermal feature of the surrounding Gaussian at index $u$ in the local neighbor set $U$, where $|U|$ denotes its cardinality.
Note that a higher $q$ increases the influence of the anchor Gaussian, while a lower value amplifies the effect of the surroundings.
This formulation ensures that updates to thermal features incorporate both the original anchor feature $F_{t_{i}}$ and the influence of the surrounding features $F_{t_{u}}$, resulting in a more physically consistent thermal representation.

\subsection{Learning Depth-aware Thermal Radiation}
\subsubsection{Depth-aware Thermal Radiation}
Since depth map rendering determines the exact position of each 3D Gaussian, the accuracy of the depth map highly influences rendered image quality. 
Therefore, several methods~\cite{kerbl2024hierarchical, turkulainen2024dn} utilize depth foundation models to improve geometric awareness by supervising depth map rendering. 
However, they require a preprocessing stage to generate depth GT.
Instead, we enforce geometric constraints for depth maps by utilizing temperature-depth relations.
Firstly, we define a depth-aware thermal radiation that is derived by leveraging the Stefan-Boltzmann law~\cite{stefan1879uber} and the inverse square law~\cite{adelberger2003tests}, as follows:
\begin{equation}
E_r = \frac{\tau \cdot t_{GT}^4}{{D_{t}}^2},
\label{eq:stefan_boltzmann}
\end{equation}
where $E_r$ represents the estimated thermal radiation, $\tau$ is the Stefan-Boltzmann constant, $t_{GT}$ denotes the temperature from GT thermal image, and $D_{t}$ is the rendered thermal depth map.
While temperature $t_{GT}$ solely represents the intrinsic thermal state of an object, thermal radiation $E_r$ accounts for both the object's temperature and its spatial distribution relative to the observer.
Since infrared sensors detect radiation rather than temperature, depth-aware radiation modeling ensures more physically consistent supervision for thermal 3D rendering.

\begin{table*}[hbt!]
\centering
\resizebox{\textwidth}{!}{
\def\arraystretch{1.35}
\begin{tabular}{lcccccccccccccccccccc}
\Xhline{3\arrayrulewidth}
\multirow{3}{*}{Methods} & \multicolumn{4}{c}{\textit{Building (Spring)}} & \multicolumn{4}{c}{\textit{Building (Winter)}}& \multicolumn{4}{c}{\textit{Exhibition}} & \multicolumn{4}{c}{\textit{Trees}} & \multicolumn{4}{c}{\textit{Double Robot}} \\
\cmidrule(lr){2-5} \cmidrule(lr){6-9} \cmidrule(lr){10-13} \cmidrule(lr){14-17} \cmidrule(lr){18-21}
& PSNR & SSIM & MAE & \# GS & PSNR & SSIM & MAE & \# GS & PSNR & SSIM & MAE & \# GS & PSNR & SSIM & MAE &  \# GS & PSNR & SSIM & MAE & \# GS \\
& $\uparrow$ & $\uparrow$ & ($^{\circ}\text{C}$ )~$\downarrow$ & (K) & $\uparrow$ & $\uparrow$ & ($^{\circ}\text{C}$ )~$\downarrow$ & (K) & $\uparrow$ & $\uparrow$ & ($^{\circ}\text{C}$ )~$\downarrow$ & (K) & $\uparrow$ & $\uparrow$ & ($^{\circ}\text{C}$ )~$\downarrow$ & (K) & $\uparrow$ & $\uparrow$ & ($^{\circ}\text{C}$ )~$\downarrow$ & (K) \\
\midrule
ThermoNeRF$^{\dagger}$~\cite{hassan2024thermonerf} & 24.25 & 0.909 & 3.383 & - & 29.09 & 0.891 & 0.728 & - & 34.79 & 0.967 & 0.333 & - & 28.63 & 0.927 & 0.385 & - & 28.25 & 0.901 & 0.465 & - \\
$\text{3DGS}_{\text{RGB-T}}$~\cite{kerbl20233d} & 27.65 & 0.978 & 2.194 & 365 & 27.77 & 0.954 & 0.743 & 363 & 31.59 & 0.980 & 0.313 & 1,100 & 33.30 & 0.971 & 0.227 & 480 & 30.92 & 0.962 & 0.316 & 487 \\
\hdashline
MrGS (Ours) & \textbf{29.46} & \textbf{0.980} & \textbf{1.574} & \textbf{193} & \textbf{30.02} & \textbf{0.966} & \textbf{0.550} & \textbf{155} & \textbf{34.81} & \textbf{0.986} & \textbf{0.263} & \textbf{508} & \textbf{33.58} & \textbf{0.974} & \textbf{0.218} & \textbf{235} & \textbf{32.06} & \textbf{0.964} & \textbf{0.319} & \textbf{232} \\
\Xhline{3\arrayrulewidth}
\end{tabular}
}
\resizebox{\textwidth}{!}{
\def\arraystretch{1.35}
\begin{tabular}{lcccccccccccccccccccc}
\multirow{3}{*}{Methods} & \multicolumn{4}{c}{\textit{Hot Water Kettle}} & \multicolumn{4}{c}{\textit{Hot Water Cup}}& \multicolumn{4}{c}{\textit{Melting Ice Cup}} & \multicolumn{4}{c}{\textit{Freezing Ice Cup}} & \multicolumn{4}{c}{\textit{Raspberry Pi}} \\
\cmidrule(lr){2-5} \cmidrule(lr){6-9} \cmidrule(lr){10-13} \cmidrule(lr){14-17} \cmidrule(lr){18-21}
& PSNR & SSIM & MAE & \# GS & PSNR & SSIM & MAE & \# GS & PSNR & SSIM & MAE & \# GS & PSNR & SSIM & MAE &  \# GS & PSNR & SSIM & MAE & \# GS \\
& $\uparrow$ & $\uparrow$ & ($^{\circ}\text{C}$ )~$\downarrow$ & (K) & $\uparrow$ & $\uparrow$ & ($^{\circ}\text{C}$ )~$\downarrow$ & (K) & $\uparrow$ & $\uparrow$ & ($^{\circ}\text{C}$ )~$\downarrow$ & (K) & $\uparrow$ & $\uparrow$ & ($^{\circ}\text{C}$ )~$\downarrow$ & (K) & $\uparrow$ & $\uparrow$ & ($^{\circ}\text{C}$ )~$\downarrow$ & (K) \\
\midrule
ThermoNeRF$^{\dagger}$~\cite{hassan2024thermonerf} & 30.97 & 0.908 & 1.000 & - & 30.05 & 0.872 & 0.787 & - & \textbf{32.24} & 0.980 & 0.248 & - & 25.35 & 0.978 & 0.958 & - & 31.01 & 0.930 & 0.448 & - \\
$\text{3DGS}_{\text{RGB-T}}$~\cite{kerbl20233d} & \textbf{35.09} & \textbf{0.976} & \textbf{0.471} & 387 & 29.07 & 0.936 & 0.558 & 574 & 31.00 & 0.986 & 0.233 & 388 & 29.76 & 0.990 & 0.467 & 138 & \textbf{37.38} & \textbf{0.989} & \textbf{0.139} & 406 \\
\hdashline
MrGS (Ours) & 34.66 & 0.971 & 0.582 & \textbf{48} & \textbf{31.44} & \textbf{0.954} & \textbf{0.393} & \textbf{47} & 31.09 & \textbf{0.987} & \textbf{0.223} & \textbf{229} & \textbf{31.93} & \textbf{0.993} & \textbf{0.377} & \textbf{108} & 37.17 & \textbf{0.989} & 0.145 & \textbf{189} \\
\Xhline{3\arrayrulewidth}
\end{tabular}
}
\caption{\textbf{Quantitative results of Thermal novel view synthesis on the ThermoNeRF dataset.} The \(\dagger\) indicates performance reproduced using the official code, while RGB-T refers to a modified version as a multi-modal framework. \textbf{Bold} means the best performance.
}
\label{table:thermal_table}
\end{table*}

\subsubsection{Uncertainty-aware S-SSIM}
After that, we utilize a structure-only SSIM (S-SSIM) to enforce structural consistency between the depth-aware radiation map $E_r$ and the rendered thermal image $I_t$, formulated as follows:
\begin{equation}
\text{S-SSIM}(I_t, E_r) = \frac{2 C_{I_tE_r}}{C_{I_t} + C_{E_r}},
\end{equation}
where \( C_{I_t} \) and \( C_{E_r} \) denote the local variances of \( I_t \) and \( E_r \), respectively, and \( C_{I_tE_r} \) represents their local covariance. 

Additionally, we enhance this physics-based supervision by introducing depth uncertainty. Since depth maps are learned by rendering pipelines without GT depth, certain regions exhibit higher uncertainty, leading to ambiguous Gaussian blending artifacts. We estimate an uncertainty value $u(D_{t})$ as follows: 
\begin{equation}
u(D_{t}) = \sigma(\psi(D_{t})),
\end{equation}
where \( \psi(\cdot) \) represents a 3$\times$3 convolutional layer. This uncertainty is then incorporated into our S-SSIM loss. The modified loss formulation is expressed as follows:
\begin{equation}
L_{boltz} =  \frac{1 - \text{S-SSIM}(I_t, E_{r})}{\mathrm{exp}(u(D_{t}))} + u(D_{t}),
\end{equation}
where \( L_{boltz} \) represents the uncertainty-aware S-SSIM loss.
By incorporating depth-aware thermal radiation as a structural constraint, our method encourages the network to focus more on reliable depth regions while reducing the influence of uncertain areas.

\subsection{Total Loss}
We employ both RGB and thermal losses to optimize both modalities, ensuring their proper convergence. We basically apply the standard $\ell_1$ and Differential-SSIM (D-SSIM) losses of 3DGS~\cite{kerbl20233d} to both modalities as follows:
\begin{equation}
L_{m} = (1 - \lambda_{D-SSIM}) L_{1}^{m} +
                 \lambda_{D-SSIM} L_{D-SSIM}^{m},
\end{equation}
where \( m \in \{c, t\}\) denotes the usage of modalities.
To further enhance the quality of thermal rendering, we incorporate edge-aware smoothness loss~\cite{turkulainen2024dn}, which encourages spatial smoothness by penalizing abrupt intensity changes in rendered images.
Given rendering result \( I_m \) and \( D_m \), where $m$ indicates modality, the smoothness loss is defined as follows:
\begin{gather}
    L_{smooth}^{m} = \sum_{x'} \left|\nabla I_m(x') \right| + \sum_{x'} \left| \nabla D_m(x') \right|, \label{eq:smooth_sepa} \\
    L_{smooth} = L_{smooth}^{c} + L_{smooth}^{t}, \label{eq:smooth_total}
\end{gather}
where $\nabla$ is the first-order differential operator along spatial directions and $\beta$ is a scaling factor. Consequently, the final loss is defined as follows:  
\begin{equation}
    L_{total} = L_{c} + L_{t} + \lambda_{smooth}L_{smooth} +\lambda_{boltz} L_{boltz}.
    \label{eq:final_loss}
\end{equation}

\begin{table*}[hbt!]
\centering
\resizebox{\textwidth}{!}{
\def\arraystretch{1.35}
\begin{tabular}{lcccccccccccccccccccc}
\Xhline{3\arrayrulewidth}
\multirow{3}{*}{Methods} & \multicolumn{4}{c}{\textit{Building (Spring)}} & \multicolumn{4}{c}{\textit{Building (Winter)}}& \multicolumn{4}{c}{\textit{Exhibition}} & \multicolumn{4}{c}{\textit{Trees}} & \multicolumn{4}{c}{\textit{Double Robot}} \\
\cmidrule(lr){2-5} \cmidrule(lr){6-9} \cmidrule(lr){10-13} \cmidrule(lr){14-17} \cmidrule(lr){18-21}
& PSNR & SSIM & LPIPS & \# GS & PSNR & SSIM & LPIPS & \# GS & PSNR & SSIM & LPIPS & \# GS & PSNR & SSIM & LPIPS &  \# GS & PSNR & SSIM & LPIPS & \# GS \\
& $\uparrow$ & $\uparrow$ & $\downarrow$ & (K) & $\uparrow$ & $\uparrow$ & $\downarrow$ & (K) & $\uparrow$ & $\uparrow$ & $\downarrow$ & (K) & $\uparrow$ & $\uparrow$ & $\downarrow$ & (K) & $\uparrow$ & $\uparrow$ & $\downarrow$ & (K) \\
\midrule
ThermoNeRF$^{\dagger}$~\cite{hassan2024thermonerf} & 19.11 & 0.624 & 0.449 & - & \textbf{19.15} & 0.597 & \textbf{0.479} & - & 20.85 & 0.570 & 0.348 & - & 17.53 & 0.567 & \textbf{0.444} & - & 17.21 & 0.604 & 0.434 & - \\
$\text{3DGS}_{\text{RGB-T}}$~\cite{kerbl20233d} & 23.17 & 0.819 & 0.320 & 365 & 17.77 & 0.603 & 0.548 & 363 & 23.42 & 0.724 & 0.303 & 1,100 & 19.22 & 0.609 & 0.501 & 480 &  20.53 & 0.759 & 0.370 & 487 \\
\hdashline
MrGS (Ours) & \textbf{24.28} & \textbf{0.831} & \textbf{0.315} & \textbf{193} & 18.64 & \textbf{0.630} & 0.520 & \textbf{155} & \textbf{23.85} & \textbf{0.731} & \textbf{0.299} & \textbf{508} & \textbf{19.47} & \textbf{0.621} &  0.472 & \textbf{235} & \textbf{22.40} & \textbf{0.807} & \textbf{0.335} & \textbf{232} \\
\Xhline{3\arrayrulewidth}
\end{tabular}
}

\centering
\resizebox{\textwidth}{!}{
\def\arraystretch{1.35}
\begin{tabular}{lcccccccccccccccccccc}
\multirow{3}{*}{Methods} & \multicolumn{4}{c}{\textit{Hot Water Kettle}} & \multicolumn{4}{c}{\textit{Hot Water Cup}}& \multicolumn{4}{c}{\textit{Melting Ice Cup}} & \multicolumn{4}{c}{\textit{Freezing Ice Cup}} & \multicolumn{4}{c}{\textit{Raspberry Pi}} \\
\cmidrule(lr){2-5} \cmidrule(lr){6-9} \cmidrule(lr){10-13} \cmidrule(lr){14-17} \cmidrule(lr){18-21}
& PSNR & SSIM & LPIPS & \# GS & PSNR & SSIM & LPIPS & \# GS & PSNR & SSIM & LPIPS & \# GS & PSNR & SSIM & LPIPS &  \# GS & PSNR & SSIM & LPIPS & \# GS \\
& $\uparrow$ & $\uparrow$ & $\downarrow$ & (K) & $\uparrow$ & $\uparrow$ & $\downarrow$ & (K) & $\uparrow$ & $\uparrow$ & $\downarrow$ & (K) & $\uparrow$ & $\uparrow$ & $\downarrow$ & (K) & $\uparrow$ & $\uparrow$ & $\downarrow$ & (K) \\
\midrule
ThermoNeRF$^{\dagger}$~\cite{hassan2024thermonerf} & 19.38 & 0.510 & \textbf{0.420} & - & 16.61 & 0.464 & \textbf{0.421} & - & \textbf{20.73} & \textbf{0.656} & \textbf{0.292} & - & 23.36 & 0.836 & 0.544 & - & 20.87 & 0.772 & 0.309 & - \\
$\text{3DGS}_{\text{RGB-T}}$~\cite{kerbl20233d} & \textbf{24.83} & \textbf{0.755} & 0.548 & 387 & 17.29 & 0.520 & 0.496 & 574 & 17.43 & 0.522 & 0.465 & 388 & 26.74 & 0.865 & 0.492 & 138 & 17.12 & 0.516 & 0.470 & 406 \\
\hdashline
MrGS (Ours) & 24.38 & 0.649 & 0.515 & \textbf{48} & \textbf{17.80} & \textbf{0.522} & 0.583 & \textbf{47} & 17.50 & 0.504 & 0.438 & \textbf{229} & \textbf{27.25} & \textbf{0.867} & \textbf{0.479} & \textbf{108} & \textbf{25.25} & \textbf{0.882} & \textbf{0.299} & \textbf{189} \\
\Xhline{3\arrayrulewidth}
\end{tabular}
}
\caption{\textbf{Quantitative results of the RGB novel view synthesis on the ThermoNeRF dataset.}}
\label{table:rgb_table}
\end{table*}

\begin{figure*}[hbt!]
    \centering
    \renewcommand{\arraystretch}{0.2}
    \begin{tabular}{@{}c@{\hskip 0.003\linewidth}c@{\hskip 0.003\linewidth}c@{\hskip 0.003\linewidth}c@{\hskip 0.003\linewidth}c@{\hskip 0.003\linewidth}c@{\hskip 0.003\linewidth}c@{\hskip 0.003\linewidth}c@{\hskip 0.003\linewidth}c}
     & \footnotesize GT RGB & \footnotesize ThermoNeRF~\cite{hassan2024thermonerf} & \footnotesize $\text{3DGS}_{\text{RGB-T}}$~\cite{kerbl20233d} & \footnotesize MrGS (ours) & \footnotesize GT Thermal & \footnotesize ThermoNeRF~\cite{hassan2024thermonerf} & \footnotesize $\text{3DGS}_{\text{RGB-T}}$~\cite{kerbl20233d} & \footnotesize MrGS (ours) \\
\\
    \rotatebox{90}{\footnotesize{\quad ~~~ \textit{Double Robot}}} & 
    \includegraphics[width=0.12\linewidth]{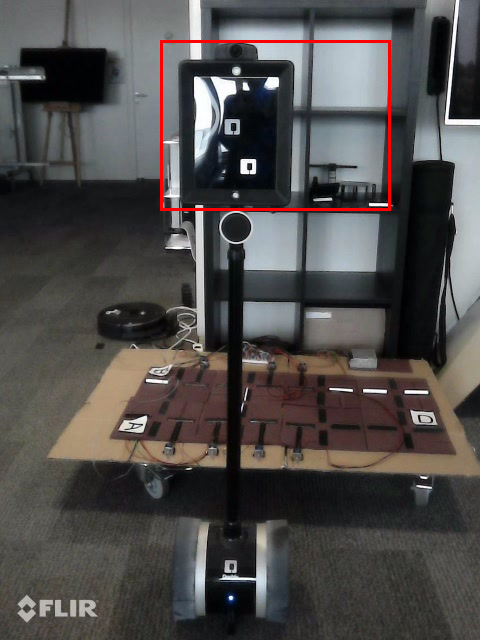} & 
    \includegraphics[width=0.12\linewidth]{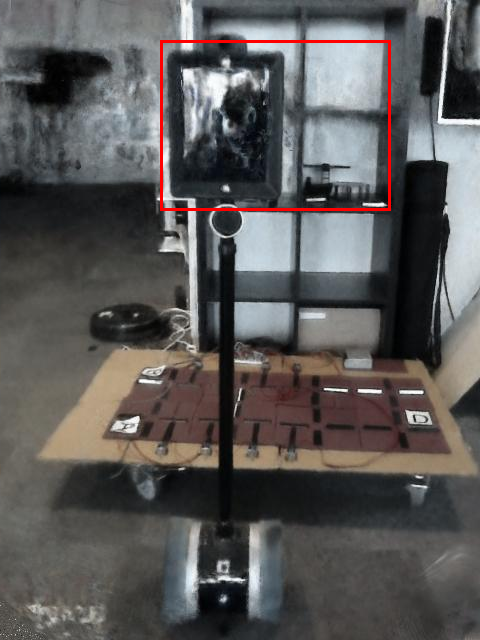} &
    \includegraphics[width=0.12\linewidth]{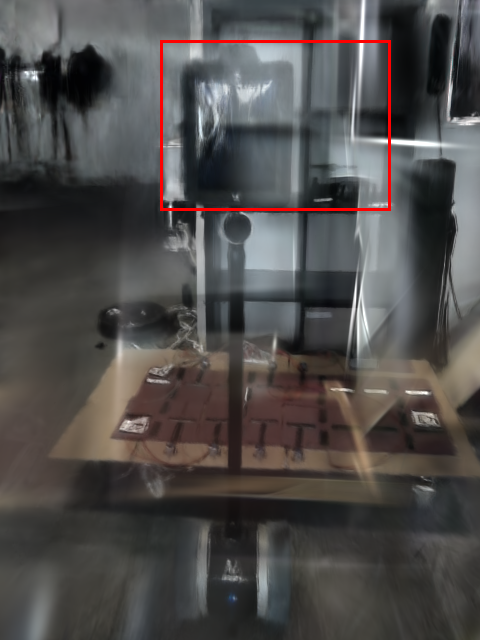} &
    \includegraphics[width=0.12\linewidth]{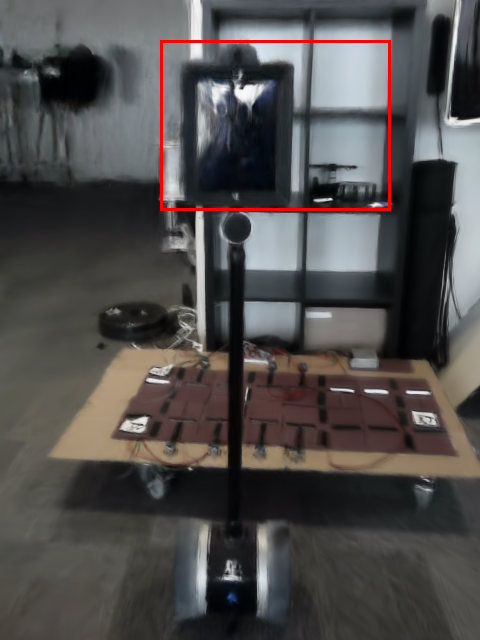} &
    \includegraphics[width=0.12\linewidth]{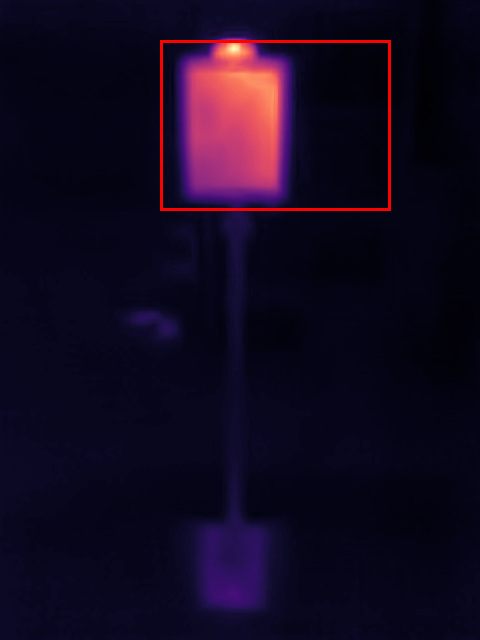} &
    \includegraphics[width=0.12\linewidth]{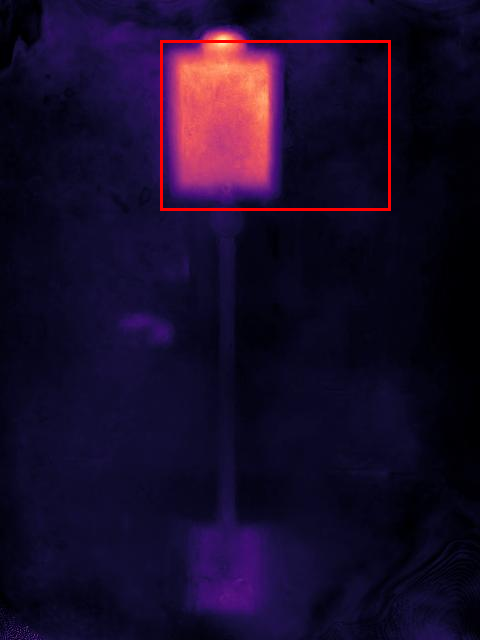} &
    \includegraphics[width=0.12\linewidth]{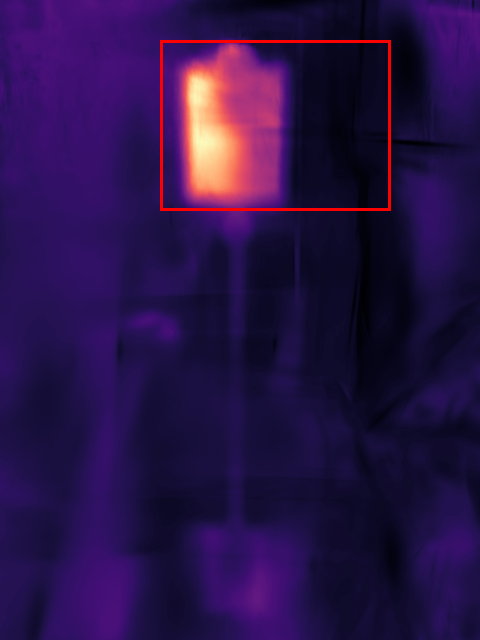} &
    \includegraphics[width=0.12\linewidth]{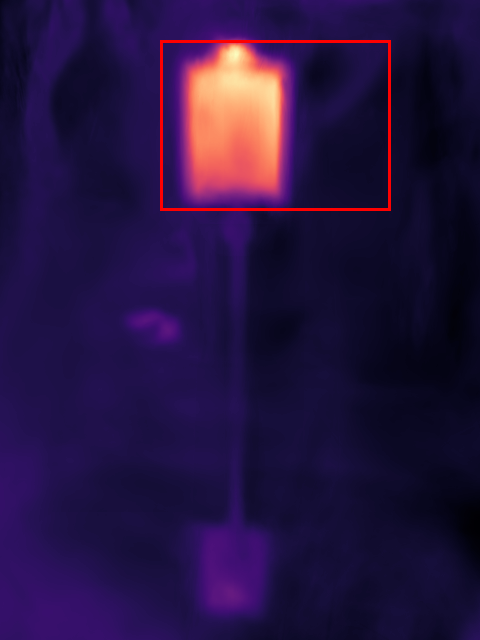}
   \\
   \rotatebox{90}{\footnotesize{\quad ~~ \textit{Hot Water Kettle}}} & 
    \includegraphics[width=0.12\linewidth]{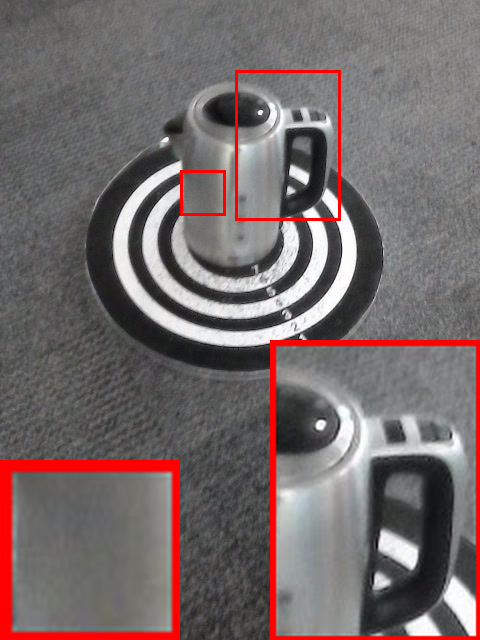} & 
    \includegraphics[width=0.12\linewidth]{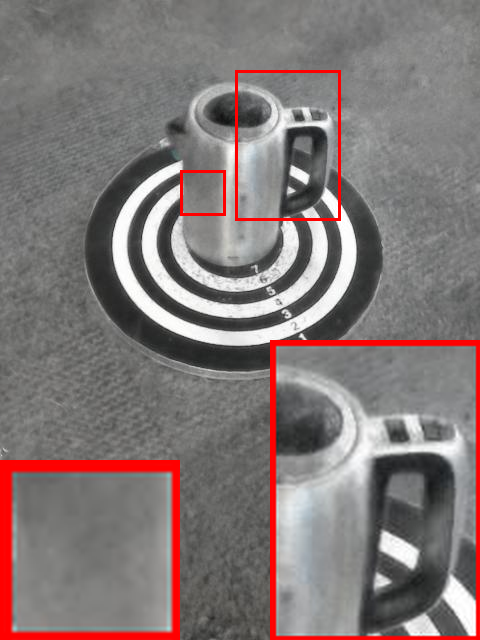} &
    \includegraphics[width=0.12\linewidth]{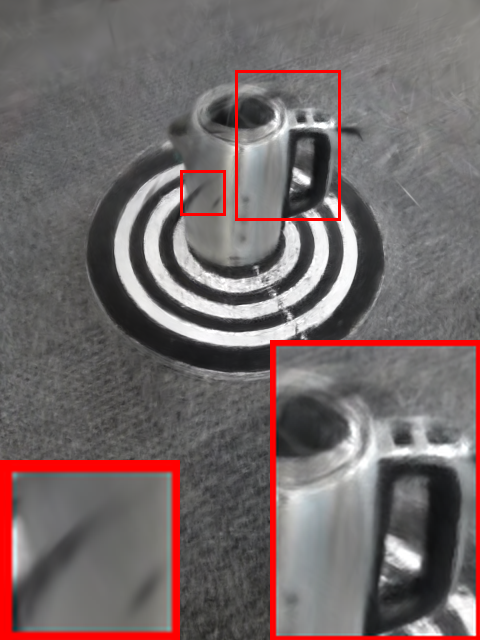} &
    \includegraphics[width=0.12\linewidth]{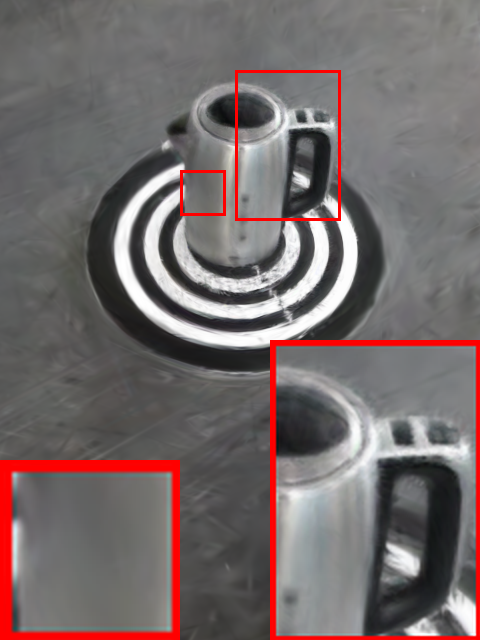} &
    \includegraphics[width=0.12\linewidth]{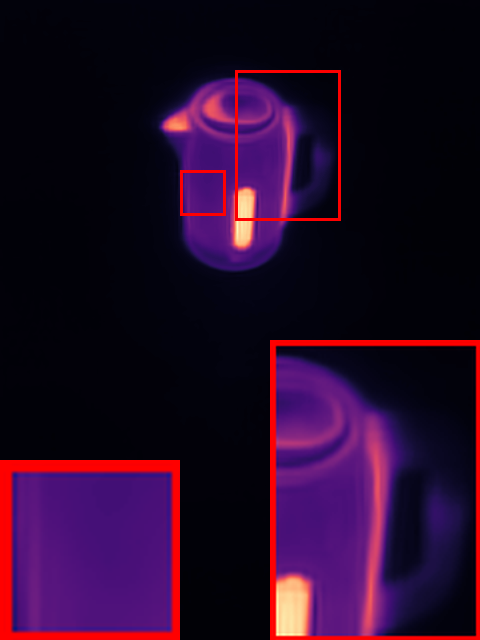} &
    \includegraphics[width=0.12\linewidth]{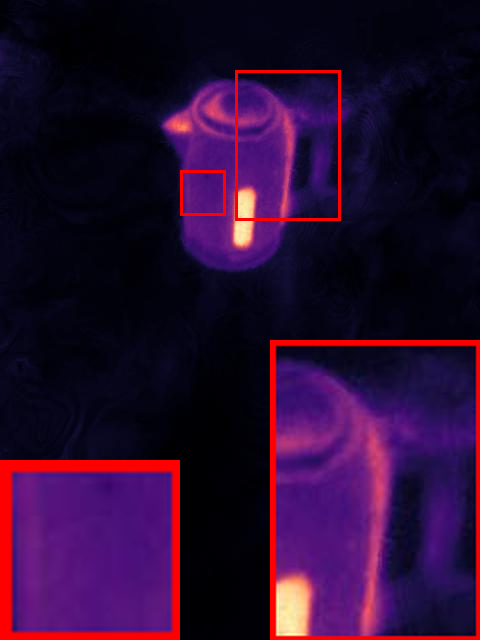} &
    \includegraphics[width=0.12\linewidth]{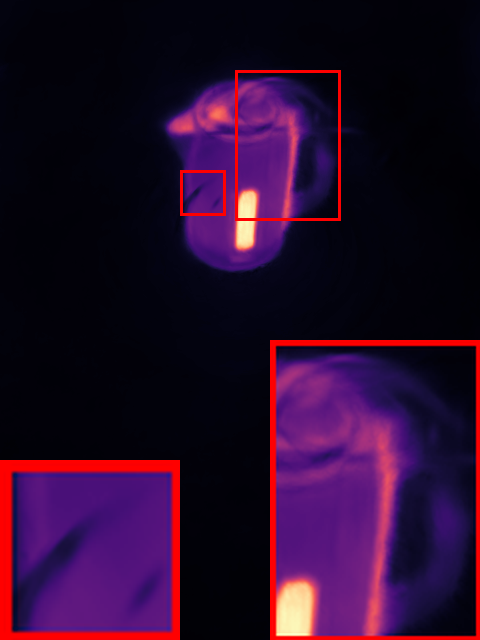} &
    \includegraphics[width=0.12\linewidth]{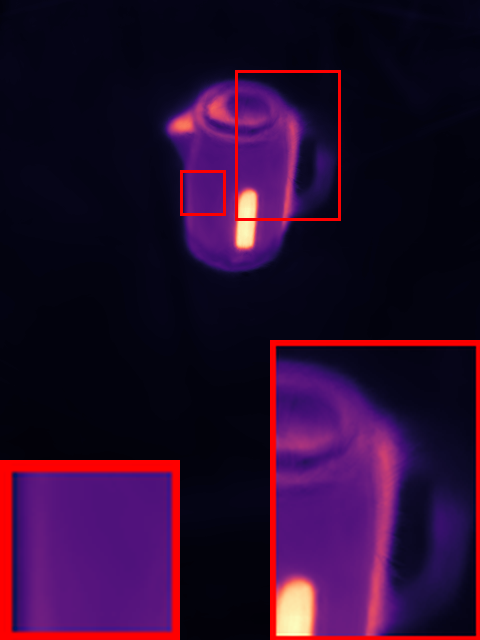}
   \\
   \rotatebox{90}{\footnotesize{\quad ~~ \textit{Freezing Ice Cup}}} & 
    \includegraphics[width=0.12\linewidth]{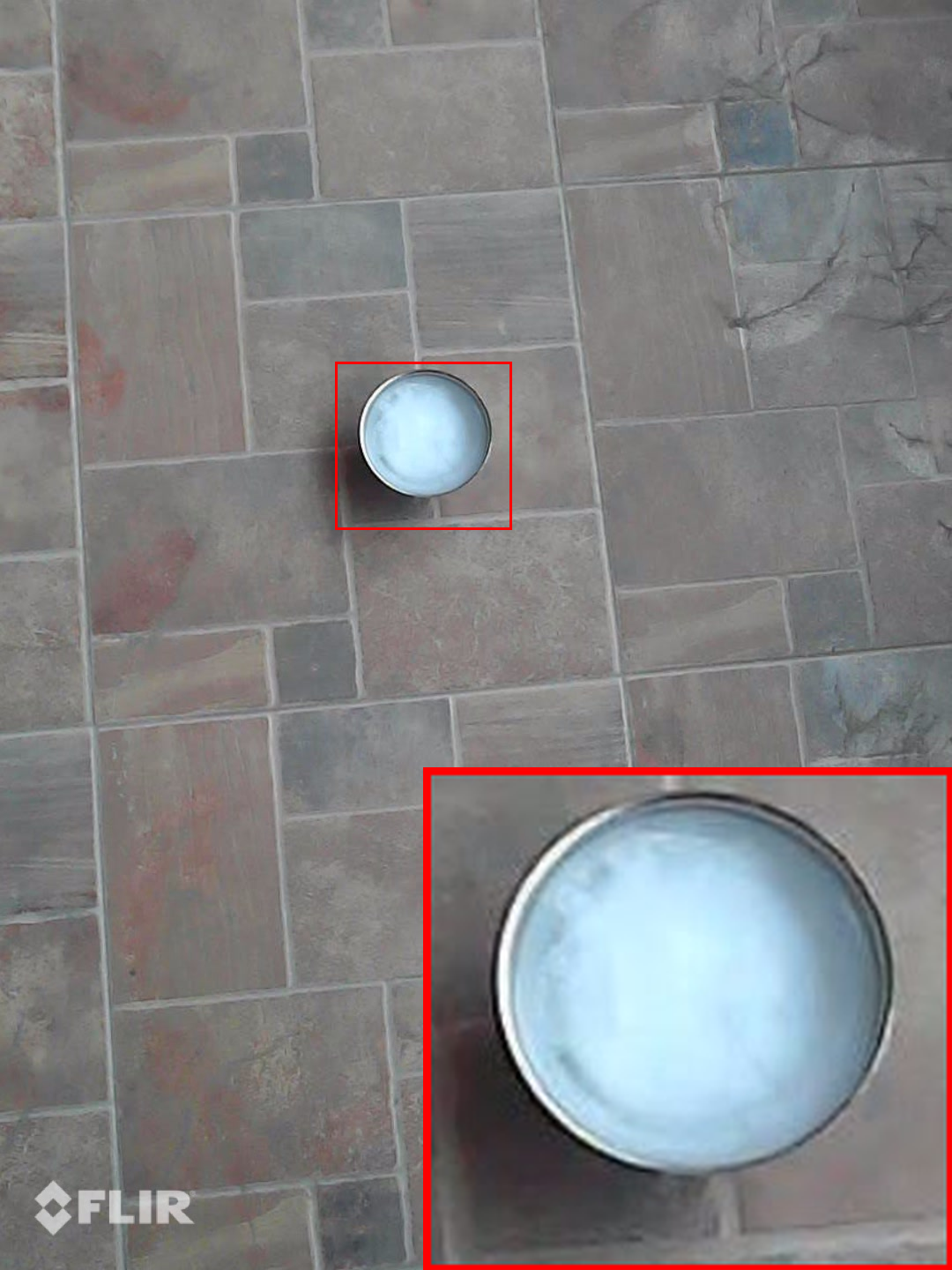} & 
    \includegraphics[width=0.12\linewidth]{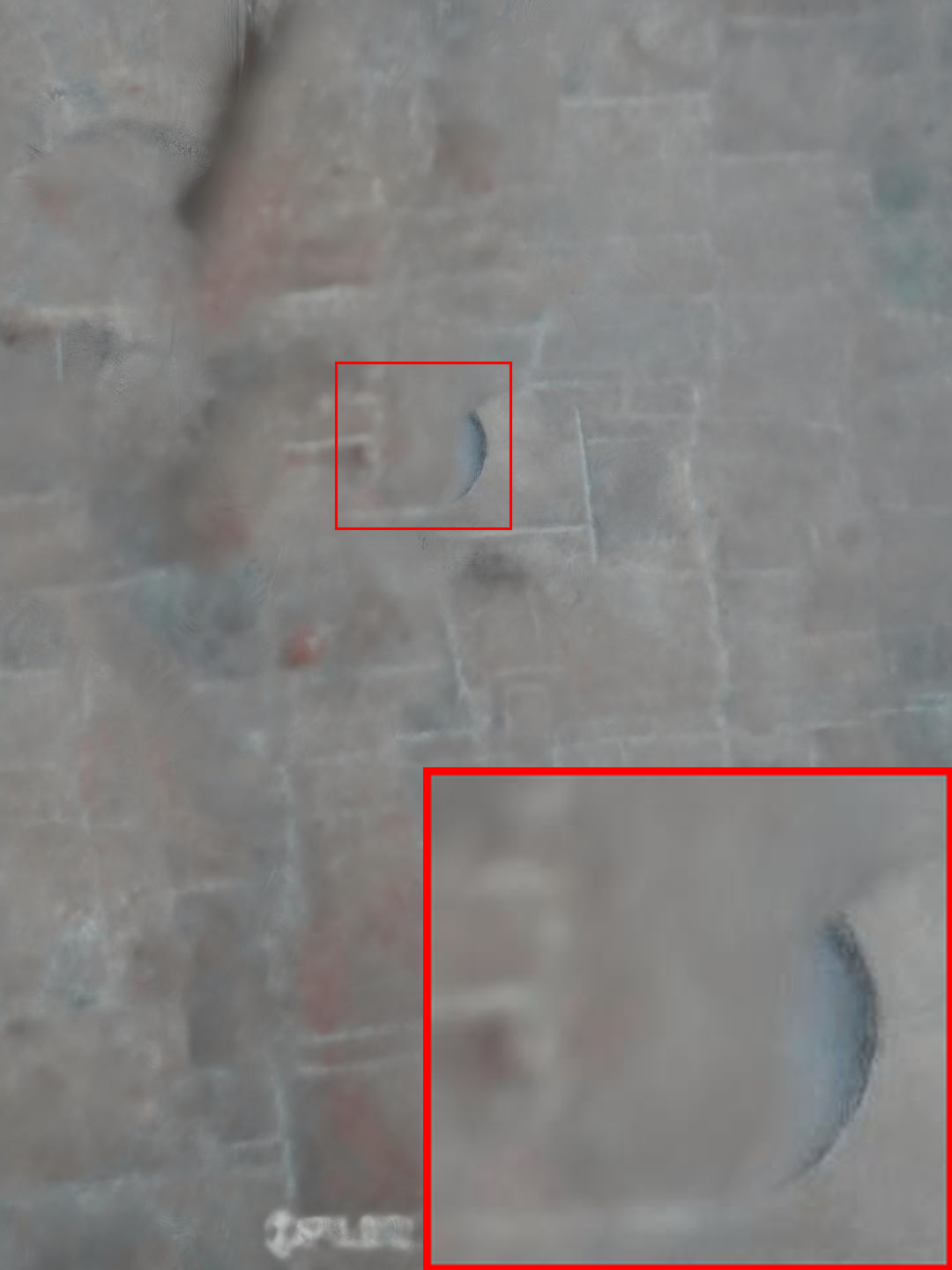} &
    \includegraphics[width=0.12\linewidth]{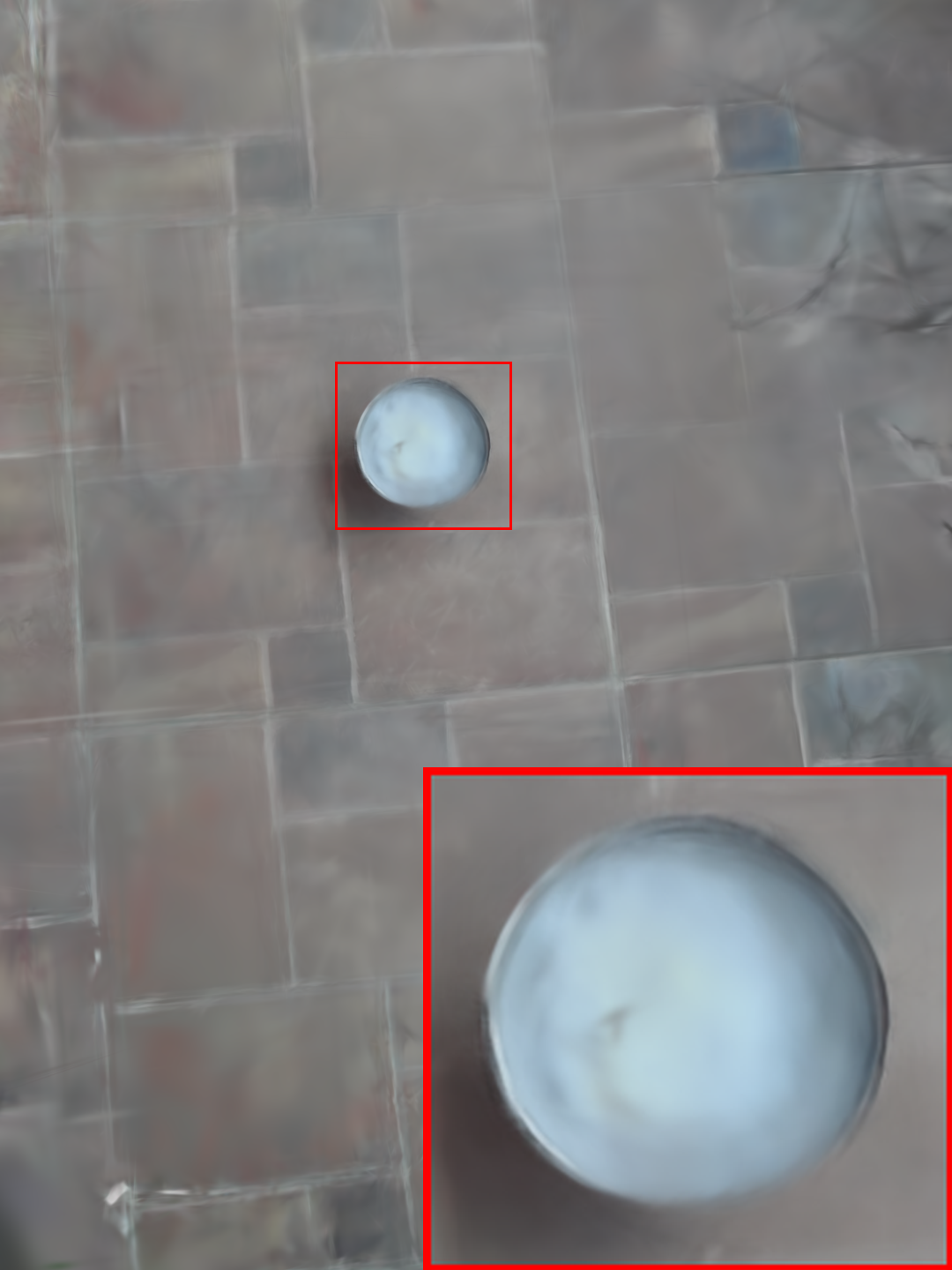} &
    \includegraphics[width=0.12\linewidth]{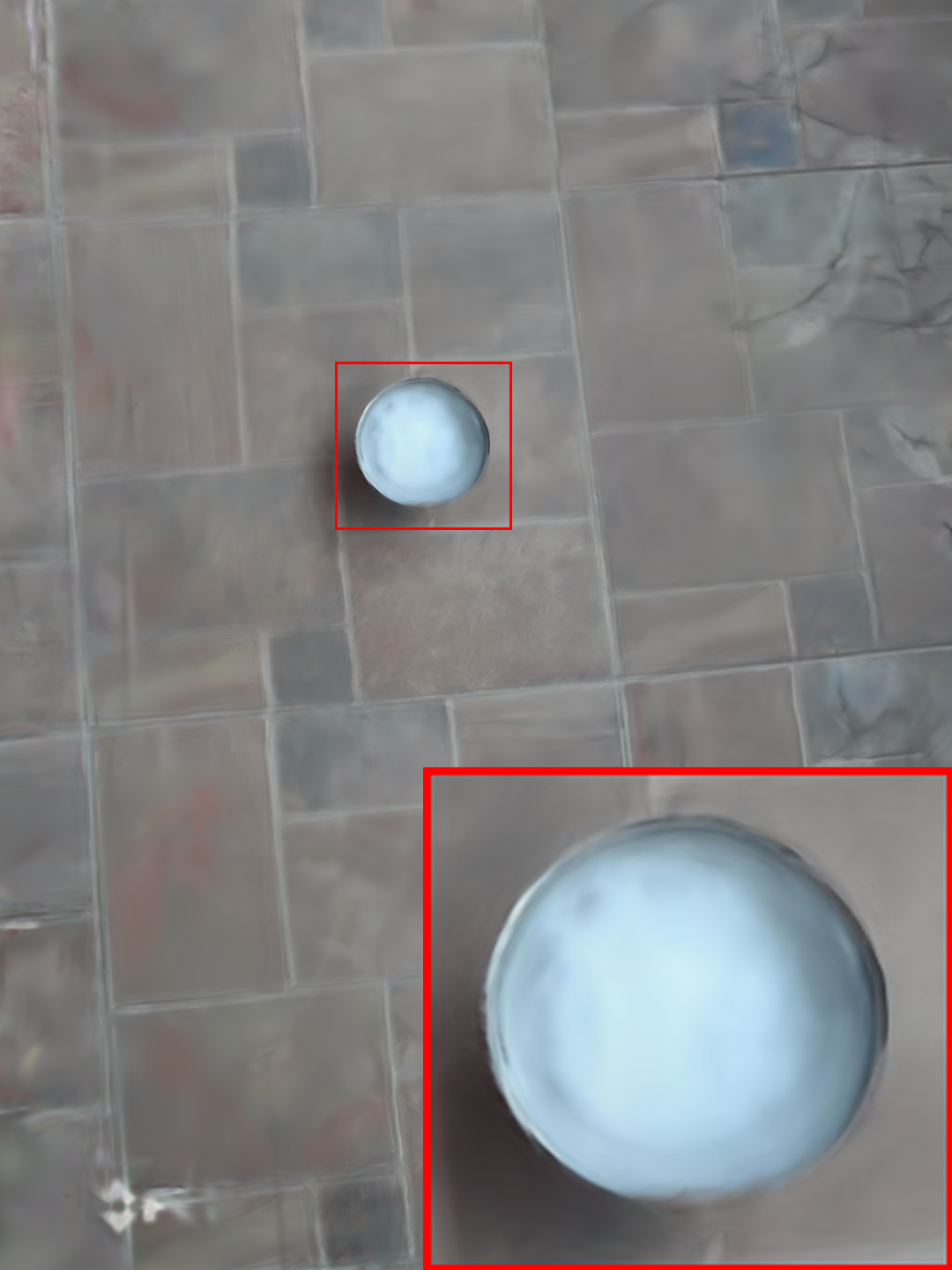} &
    \includegraphics[width=0.12\linewidth]{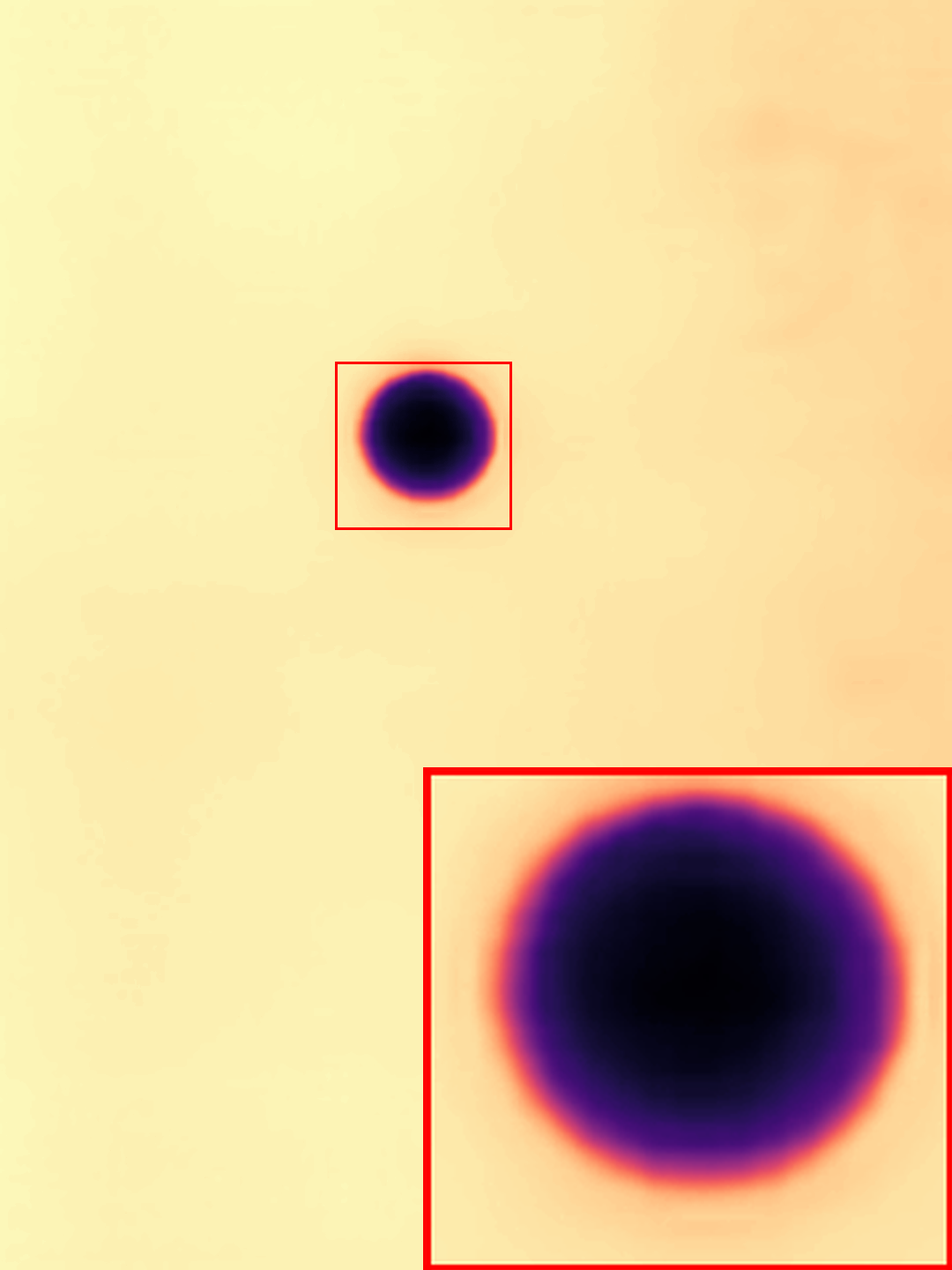} &
    \includegraphics[width=0.12\linewidth]{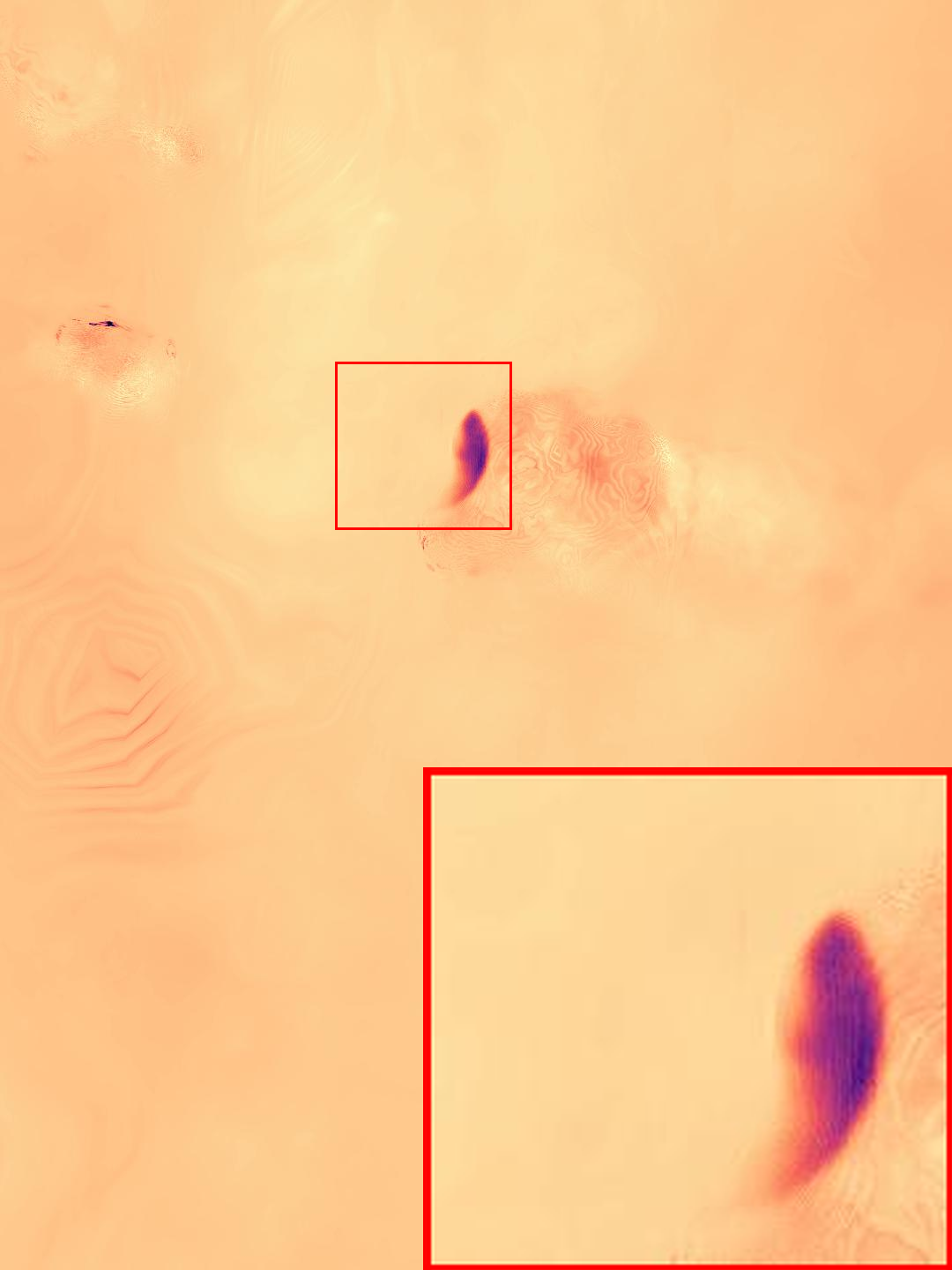} &
    \includegraphics[width=0.12\linewidth]{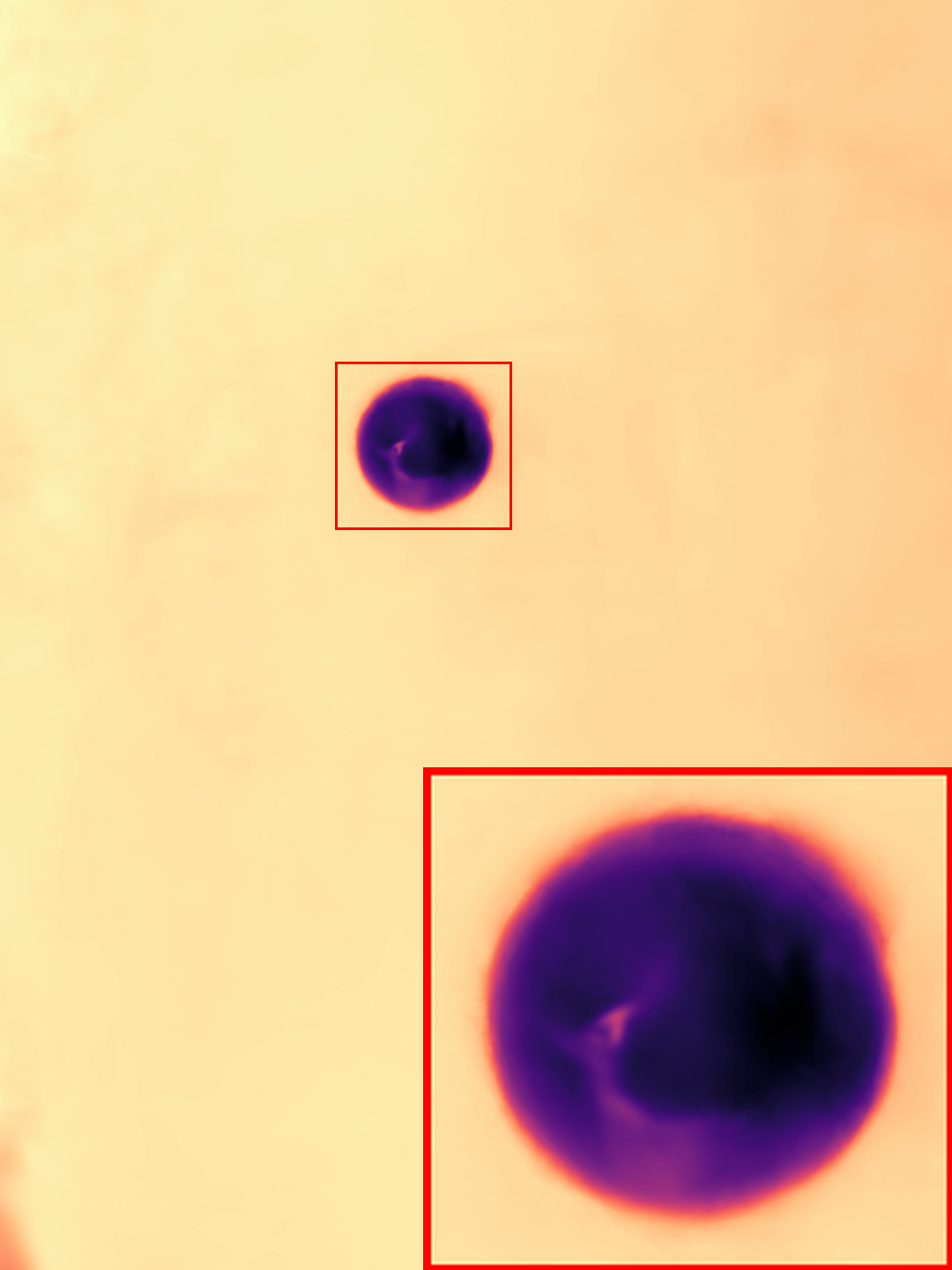} &
    \includegraphics[width=0.12\linewidth]{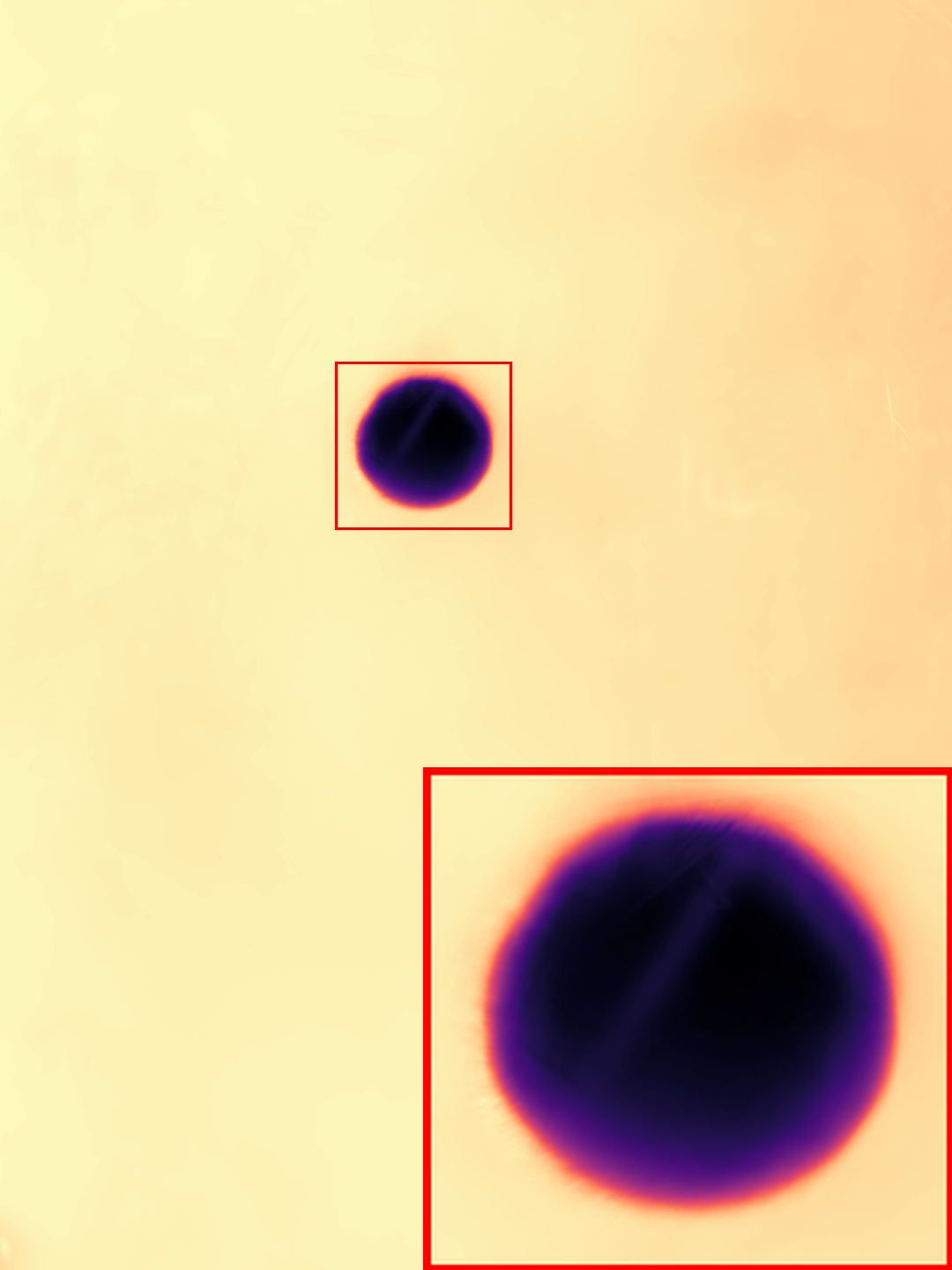}
   \\
   
    \end{tabular}
    \caption{
    \textbf{Qualitative highlights for RGB-T novel view synthesis.}
    }
\label{fig:Qualitative_Results}
\end{figure*}

\section{Experiment}
\subsection{Dataset and Metrics}
We evaluate our method on the ThermoNeRF dataset~\cite{hassan2024thermonerf}, which includes a total of 10 released RGB-T scenes. These comprise four outdoor scenes \{\textit{Building(Spring)}, \textit{Building(Winter)}, \textit{Exhibition}, \textit{Trees}\} and the remaining indoor instance scenes. The dataset was collected using a FLIR One Pro LT camera, capturing aligned RGB and thermal image pairs with a temperature range of \SI{-20}{\celsius} to \SI{120}{\celsius}.
During training and evaluation, we adopt the original image partitioning as provided in the official split. We evaluate the performance of our method using PSNR, SSIM~\cite{wang2004image}, and MAE (\SI{}{\celsius})~\cite{hassan2024thermonerf} for thermal novel view synthesis, and PSNR, SSIM, and LPIPS~\cite{zhang2018unreasonable} for RGB novel view synthesis. For LPIPS, we use VGG~\cite{simonyan2014very} network to evaluate perceptual similarity.

\subsection{Implementation Details}
\subsubsection{MrGS}
For the loss weights, we set \(\lambda_{D-SSIM}\) and \(\lambda_{smooth}\) to 0.2 and 0.6 respectively. \(\lambda_{boltz}\) was decayed from 0.05 to 0.01 during the densification period. All experiments were conducted on an RTX A6000 GPU.
Please refer to our supplementary material for more details.

\subsubsection{Modified 3DGS for RGB-T rendering}
Since there are no prior 3DGS methods that have addressed multi-modal rendering, we extend original 3DGS~\cite{kerbl20233d} into a multi-modal framework for comparison with our approach.
Specifically, we modify 3DGS by incorporating spherical harmonics for thermal rendering, enabling thermal values to be rasterized in the same manner as RGB colors. 
Through this benchmarking, we demonstrate that our framework is better suited for multi-modal rendering than simply combining existing methods, as it explicitly accounts for the distinct physical properties of each modality.

\subsection{Quantitative and Qualitative Evaluations}
\subsubsection{Thermal Novel View Synthesis}
We present the quantitative results evaluated in thermal scenes in \cref{table:thermal_table}.
Our proposed method maintains high rendering quality regardless of seasonal changes, demonstrating robustness to varying thermal conditions thanks to our physics-driven approaches.  
When we tested on four outdoor scenes \{\textit{Building(Spring)}, \textit{Building(Winter)}, \textit{Exhibition}, \textit{Trees}\}, our method produces an average PSNR that is 2.78dB higher than ThermoNeRF~\cite{lin2024thermalnerf}.
Furthermore, our MrGS achieves an average PSNR that is 1.89dB higher than the modified 3DGS, while using about 50\% fewer Gaussians.

MrGS also demonstrates superior rendering performance on the remaining indoor scenes. In the \textit{Hot Water Kettle} and \textit{Hot Water Cup} scenes, our method achieves comparable or superior rendering performance while demonstrating its remarkable efficiency in terms of the number of Gaussians.
MrGS only utilizes 12.4\% and 8.2\% of the Gaussians compared to modified 3DGS.
This reduction is driven by our framework design, which incorporates multi-modal appearance embedding, orthogonal extraction, uncertainty-aware thermal radiation, and Fourier heat transformation. \Cref{fig:Qualitative_Results} provides qualitative comparison results.
In the \textit{Hot Water Kettle} scene, our method exhibits significantly fewer floaters compared to 3DGS. 
Moreover, it accurately captures the complete temperature distribution in the \textit{Freezing Ice Cup}, outperforming both ThermoNeRF~\cite{hassan2024thermonerf} and 3DGS.

\begin{table}[t]
\centering
\resizebox{1.00\linewidth}{!}{
\renewcommand{\arraystretch}{1.5}
\begin{tabular}{c|cccc|ccc|c}
\Xhline{3\arrayrulewidth}
\multirow{2}{*}{Scenes} & \multicolumn{4}{c|}{Method} & \multicolumn{3}{c|}{Metrics} & \# GS \\ \cline{2-8}
& \text{3DGS\textsubscript{RGB-T}} & $\text{MrGS}_{base}$ & Fourier & $L_{boltz}$ & PSNR & SSIM & MAE(°C) & (K) \\ \hline\hline
\multirow{5}{*}{\makecell{\textit{Building} \\ \textit{(Winter)}}} & \checkmark &  &  &  & 27.95 & 0.957 & 0.764 & 342 \\
& & \checkmark &  &  & 29.28 & 0.963 & 0.599 & 201 \\
& & \checkmark & \checkmark & & 29.53 & \underline{0.965} & 0.585 & \underline{160} \\
& & \checkmark &  & \checkmark & \underline{29.81} & \underline{0.965} & \underline{0.564} & 188 \\
& & \checkmark & \checkmark & \checkmark & \textbf{30.02} & \textbf{0.966} & \textbf{0.550} & \textbf{155} \\ \hline
\multirow{5}{*}{\makecell{\textit{Melting} \\ \textit{Ice Cup}}} & \checkmark &  &  &  & 27.07 & 0.966 & 0.546 & 267 \\
& & \checkmark &  &  & 30.28 & 0.984 & 0.268 & 265 \\

& & \checkmark & \checkmark & & \underline{30.93} & \underline{0.986} & \underline{0.238} & 258 \\
& & \checkmark &  & \checkmark & 30.35 & 0.985 & 0.249 & \underline{228} \\
& & \checkmark & \checkmark & \checkmark & \textbf{31.09} & \textbf{0.987} & \textbf{0.223} & \textbf{224} \\
\Xhline{3\arrayrulewidth}
\multicolumn{9}{r}{\textbf{Bold}: The best, \underline{Underline}: The second-best}
\end{tabular}}
\caption{\textbf{Ablation study of our proposed methods.}
}
\label{tab:ab_combination}
\end{table}

\subsubsection{RGB Novel View Synthesis}
\Cref{table:rgb_table} provides RGB scene reconstruction performance comparisons on the ThermoNeRF dataset.
By leveraging both RGB and thermal data, our method enhances scene reconstruction performance and material awareness, enabling effective representations while maintaining high-quality rendering.
For indoor instances, our method achieves over a 20.9\% PSNR improvement compared with ThermoNeRF and 3DGS in the \textit{Raspberry Pi} scene.
Our method also demonstrates a PSNR increase of 5.19, SSIM improvement of 0.203, and LPIPS reduction of 0.099 in the \textit{Double Robot} scene compared with ThermoNeRF, showcasing our robustness in multi-modal reconstruction.

Although MrGS exhibits a slight decrease in PSNR and LPIPS compared to 3DGS in the \textit{Hot Water Kettle} scene, it maintains comparable performance while only using 12.4\% of the Gaussians.
This efficiency highlights the effectiveness of our multi-modal framework, which utilizes multi-modal appearance embedding to encompass both color and thermal modality.
Integrating thermal data with RGB further enhances rendering quality and maximizes Gaussian usage, improving both efficiency and scalability in multi-modal scenarios.
\Cref{fig:Qualitative_Results} shows a qualitative comparison of RGB rendering results.
In the \textit{Double robot} scene, our method accurately renders the image, while 3DGS exhibits a hallucination effect and ThermoNeRF suffers from blurriness.
Moreover, MrGS does not exhibit floater artifacts that can adversely affect RGB quality, as shown in the \textit{Hot Water Kettle} scene.

\begin{table}[t]
\centering
\resizebox{1.0\linewidth}{!}{
\renewcommand{\arraystretch}{1.35}
\begin{tabular}{c|ccc|cc}
\Xhline{3\arrayrulewidth}
{Sample $K$} & \multicolumn{3}{c|}{Metrics} & \# GS & Training time \\ \cline{2-4}
{($K=n^2 - 1$)} & PSNR & SSIM & MAE(°C) & (K) & (min) \\ \hline\hline
\(n=1\)& 31.24 & \underline{0.952} & \textbf{0.381} & 42 & 17 \\
\(n=3\)& \textbf{31.44} & \textbf{0.954} & \underline{0.393} & 47 & 27\\
\(n=5\)& \underline{31.39} & 0.945 & 0.394 & 47 & 35 \\
\(n=7\)& 29.84 & 0.942 & 0.489 & 47 & 58 \\
\Xhline{3\arrayrulewidth}
\multicolumn{6}{r}{\textbf{Bold}: The best, \underline{Underline}: The second-best}
\end{tabular}}
\caption{\textbf{Ablation on Fourier Heat Transformation with the number of sampled Gaussians.}
}
\label{tab:ab_fourier_n}
\end{table}
\begin{figure}[t]
    \centering
    \renewcommand{\arraystretch}{0.2}
    \begin{tabular}{c@{\hskip 0.003\linewidth}c@{\hskip 0.003\linewidth}c@{\hskip 0.003\linewidth}c}
        \includegraphics[width=0.24\linewidth]{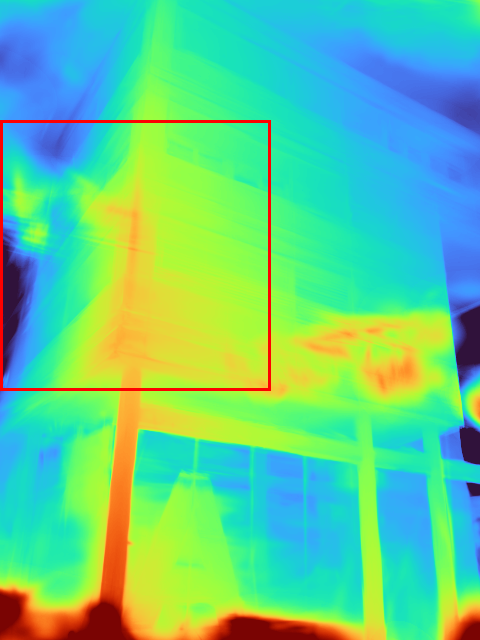}  & 
        \includegraphics[width=0.24\linewidth]{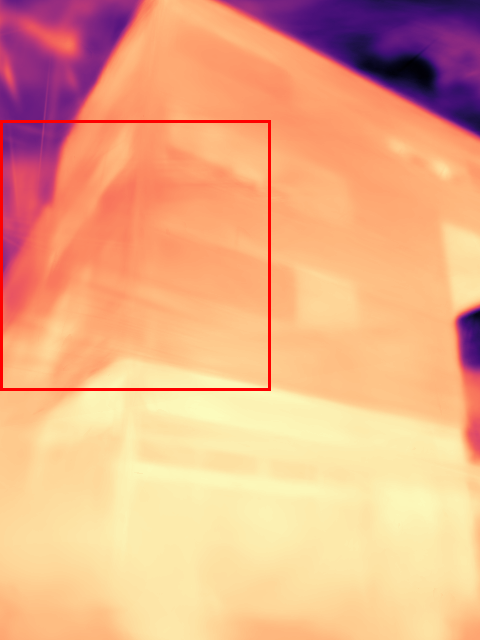} &
        \includegraphics[width=0.24\linewidth]{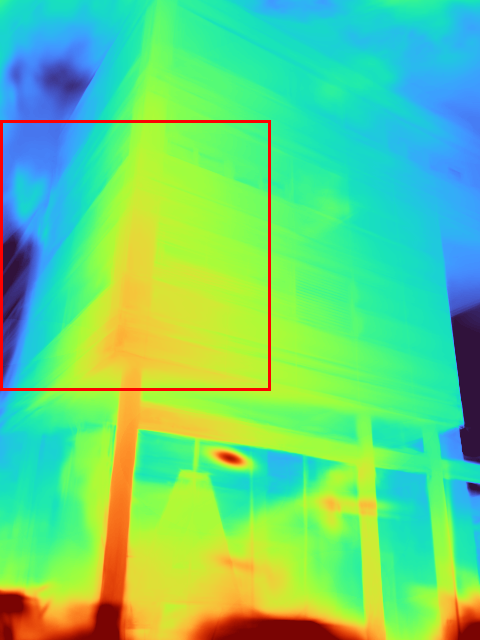}  & 
        \includegraphics[width=0.24\linewidth]{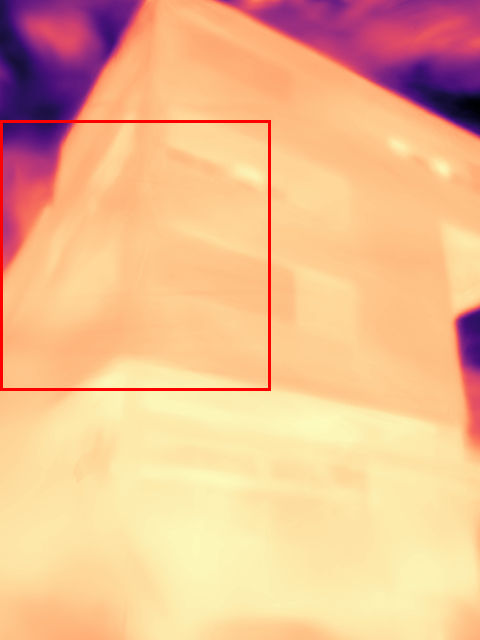} \\
        \footnotesize Depth map & \footnotesize Rendered thermal & \footnotesize Depth map & \footnotesize Rendered thermal \\
        \footnotesize  (w/o $L_{boltz}$) & \footnotesize  (w/o $L_{boltz}$) & \footnotesize  (w/ $L_{boltz}$) & \footnotesize  (w/ $L_{boltz}$)
    \end{tabular}
    \caption{\textbf{Effectiveness of the proposed depth-aware thermal radiation learning ($L_{boltz}$) with an uncertainty.}}
    \label{fig:boltzmann}
\end{figure}

\subsection{Ablation Study}

\subsubsection{Effectiveness of Our Proposed Methods}
\Tabref{tab:ab_combination} presents our ablation study on different combinations of the proposed methods, evaluating their effectiveness in the \textit{Building (Winter)} and \textit{Melting Ice Cup} scenes.
Our proposed framework (\ie, \text{MrGS\textsubscript{base}}) consistently outperforms the base architecture (\ie, \text{3DGS\textsubscript{RGB-T}}) by incorporating orthogonal extraction.
To assess the individual contributions of our proposed physical-driven techniques, we analyze performance trends when integrating Fourier heat transformation (\ie, Fourier) and depth-aware thermal radiation learning (\ie, $L_{boltz}$) into the base model.
Utilizing heat flux from Fourier heat transformation significantly improves PSNR and LPIPS, particularly in the outdoor building scene, while also reducing the number of Gaussians. This demonstrates that incorporating surrounding thermal interactions into rasterization effectively models real-world heat flow, enhancing thermal scene reconstruction.

Meanwhile, incorporating $L_{boltz}$ leads to a noticeable reduction in Gaussian count and improves evaluation metrics, especially in indoor scenes.
This suggests that integrating Stefan-Boltzmann law-based radiation modeling with structure-SSIM supervision reinforces geometric constraints in thermal rendering, even when texture alone is insufficient for inferring scene structure.
Additionally, incorporating depth information using the uncertainty-aware approach further improves structural detail preservation, reduces floaters, and enhances overall thermal rendering quality, as shown in \figref{fig:boltzmann}.
By combining both approaches, we achieve the highest performance across both instance-level indoor and large-scale outdoor scenes, demonstrating their complementary benefits.

\subsubsection{Quantitative Effect of Heat Flux Modeling}
\Tabref{tab:ab_fourier_n} shows an ablation study on the number of Gaussians sampled during the Fourier Heat Transformation's $K$-NN search (\cf \cref{sec:fourier}), tested in the \textit{Hot Water Cup} scene.
Through the proposed Fourier Heat Transformation, we enable each Gaussian to learn a temperature attribute while refining its heat distribution by considering heat flux between Gaussians before rendering. The heat flux depends on the temperature gradient, which we approximate using a convolution over the sampled Gaussians. As the number of sampled Gaussians increases, a broader range is considered for heat flux computation. 
Experimental results show that when $n = 3$, our method achieves the highest performance in two out of three rendering metrics while maintaining a reasonable training time. Therefore, we set $n = 3$ as the default configuration.

\section{Conclusion}
In this work, we introduced MrGS, the 3DGS-based multi-modal radiance field, enabling efficient RGB-T scene reconstruction. Our framework effectively disentangles modality-specific information through orthogonal feature extraction from a single appearance feature and models RGB–thermal intensity using embedding strategies that adapt appropriately to the degree of Lambertian reflectance. To improve thermal rendering, we incorporated Fourier’s law of heat conduction to simulate heat transfer between neighboring Gaussians. Furthermore, we proposed a depth-aware thermal radiation map based on the Stefan-Boltzmann law and the inverse-square law. Experimental results demonstrate that MrGS achieves high-fidelity multi-modal rendering with significantly fewer Gaussians while outperforming existing NeRF- and 3DGS-based methods across various RGB-T scenes.

{\small
\bibliographystyle{IEEEtran}
\bibliography{main}
}

\end{document}